\documentclass[fleqn,10pt]{wlscirep}
\usepackage[utf8]{inputenc}
\usepackage[T1]{fontenc}
\usepackage{multirow}
\usepackage{subcaption}
\usepackage{float}

\usepackage{threeparttable}  
\title{Detection of AI Generated Images Using Combined Uncertainty Measures and Particle Swarm Optimised Rejection Mechanism}

\author[1,+]{Rahul Yumlembam}
\author[1,*+]{Biju Issac}
\author[1,+]{Nauman Aslam}
\author[1,+]{Eaby Kollonoor Babu}
\author[2,+]{Josh Collyer}
\author[2,+]{Fraser Kennedy}

\affil[1]{Northumbria University, Department of Computer and Information Sciences, Newcastle upon Tyne, NE1 8ST, United Kingdom}
\affil[2]{The Alan Turing Institute, British Library, 96 Euston Road, London NW1 2DB, United Kingdom}

\affil[*]{bissac@ieee.org}

\affil[ ]{Emails: 
r.yumlembam@northumbria.ac.uk,
bissac@ieee.org, nauman.aslam@northumbria.ac.uk, e.k.babu@northumbria.ac.uk, jcollyer@turing.ac.uk, fkennedy@turing.ac.uk}

\begin{abstract}
As AI-generated images become increasingly photorealistic, distinguishing them from natural images poses a growing challenge. This paper presents a robust detection framework that leverages multiple uncertainty measures to decide whether to trust or reject a model's predictions. We focus on three complementary techniques: Fisher Information, which captures the sensitivity of model parameters to input variations; entropy-based uncertainty from Monte Carlo (MC) Dropout, which reflects predictive variability; and predictive variance from a Deep Kernel Learning (DKL) framework using a Gaussian Process (GP) classifier. To integrate these diverse uncertainty signals, we employ Particle Swarm Optimisation (PSO) to learn optimal weightings and determine an adaptive rejection threshold. The model is trained on Stable Diffusion-generated images and evaluated on GLIDE, VQDM, Midjourney, BigGAN and StyleGAN3 each presenting significant distribution shifts. While standard metrics like prediction probability and Fisher-based measures perform well in distribution, they degrade under shift. In contrast, the Combined Uncertainty measure consistently achieves an incorrect rejection rate of approximately 70\% on unseen generators, successfully filtering out most misclassified AI samples. Although the system occasionally rejects correct predictions from newer generators, this conservative behaviour remains acceptable, as rejected synthetic samples serve as valuable input for retraining. Crucially, it maintains high acceptance of accurate predictions for natural images and in-domain AI data.
Under adversarial attacks (FGSM and PGD), the Combined Uncertainty method rejects around 61\% of successful attacks, while the GP-based uncertainty alone achieves up to 80\%. Notably, the Combined approach maintains strong selectivity, rarely rejecting correct predictions. Overall, our findings highlight the benefit of multi-source uncertainty fusion for resilient and adaptive AI-generated image detection.
\end{abstract}
\begin{document}

\flushbottom
\maketitle
%
%
\thispagestyle{empty}


\section*{Introduction}
The rapid evolution of artificial intelligence (AI) and machine learning has not only revolutionised traditional computing paradigms but has also fundamentally transformed the field of image generation. Over the past decade, cutting-edge techniques have emerged that enable machines to produce images with astonishing realism. Models such as Generative Adversarial Networks (GANs) \textsuperscript{\cite{goodfellow2020generative},\cite{lu2018image}} have been at the forefront of this revolution, introducing adversarial frameworks that pit two neural networks against each other to refine image quality iteratively. Similarly, Diffusion models \textsuperscript{\cite{ho2020denoising}} have harnessed the power of stochastic processes to reverse-engineer high-quality images from noisy inputs. These technological breakthroughs have not only broadened the horizons of what is possible in digital content creation but have also increased rapid innovation across various creative fields. In art and design, for instance, these generative models empower artists to explore novel aesthetics and generate complex visual patterns that were previously unimaginable. In the entertainment industry, they contribute to more immersive visual effects and realistic character renderings, enhancing the overall viewer experience. However, while these advancements have catalysed creativity and innovation, they have concurrently introduced a host of challenges that extend beyond the realm of art.

One of the most pressing concerns is the potential for misuse in the spread of misinformation and digital forgery. The hyper-realistic images produced by these models can be weaponised to create deceptive content that undermines public trust in media. As synthetic images become increasingly convincing, there is a growing risk that they could be deployed to manipulate opinions, distort facts, or even influence political outcomes. This erosion of public trust is particularly alarming in a digital age where visual media often serves as a primary source of information and evidence. Traditionally, classifiers trained on comprehensive datasets containing natural and AI-generated images have performed robustly within the confines of the known data distribution. These classifiers rely on patterns learned during training to distinguish between authentic and synthetic images accurately. However, a critical vulnerability arises when novel generative techniques produce images that fall outside the scope of these training datasets. As new methods evolve rapidly, they generate images with features and characteristics that differ significantly from those encountered during training. This phenomenon, known as the out-of-distribution (OOD) problem, poses a serious threat to the reliability of conventional classifiers. When faced with OOD images, classifiers may fail to recognise subtle deviations, allowing these images to bypass detection mechanisms \textsuperscript{\cite{zhu2024genimage},\cite{cozzolino2018forensictransfer},\cite{zhang2019detecting},\cite{park2024performance}}.

Recent research has underscored the fragility of current classification models in the face of such distributional shifts. Even minor changes in the data distribution can lead to marked performance degradation, highlighting the inherent challenges in reliably detecting OOD examples \textsuperscript{\cite{hendrycks2016baseline},\cite{liang2017enhancing}}. This degradation is particularly concerning in real-world applications where the data landscape continually evolves, and adversaries may deliberately exploit these vulnerabilities. The resulting uncertainty not only compromises the integrity of automated systems but also calls into question the broader trustworthiness of AI-driven decision-making processes.

\begin{figure}[h]
    \centering
    \includegraphics[width=\linewidth]{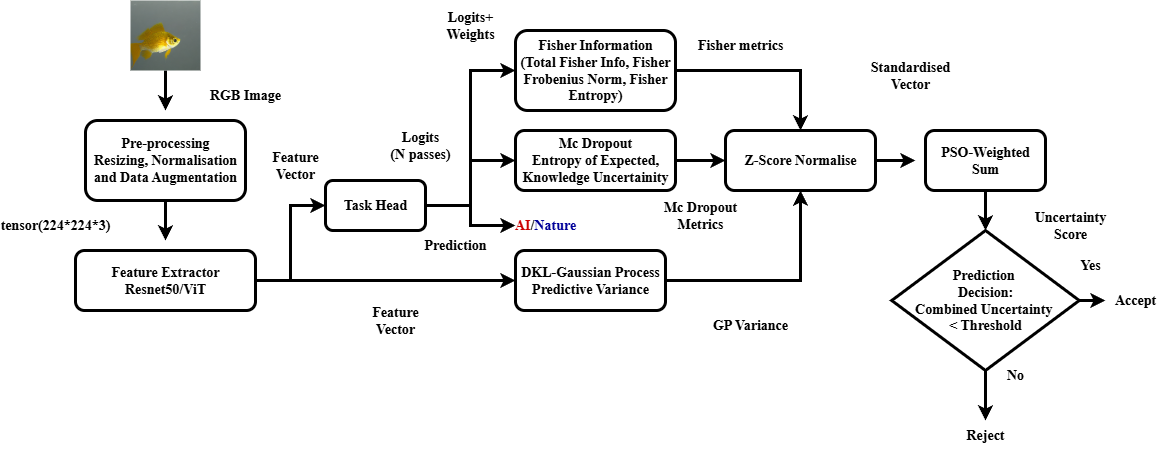}
    \caption{Overview of the proposed framework. 
    Input images are preprocessed and passed through a feature extractor and task head for classification. 
    Fisher Information, MC Dropout, and GP variance are computed in parallel, normalised, and combined via PSO to produce a unified uncertainty score for accept/reject decisions.}
    \label{fig:pipeline}
\end{figure}

This work addresses the challenge by building on existing research and introducing a unified framework that integrates complementary uncertainty measures. The framework combines entropy derived from Monte Carlo dropout, Gaussian Process predictive variance, and several Fisher Information-based metrics, namely, total Fisher information, the Frobenius norm of the Fisher Information Matrix, and Fisher entropy, to yield a single scalar uncertainty value for each sample. While MC dropout offers an ensemble-like estimate of epistemic uncertainty through multiple stochastic forward passes, its uncertainty estimates are often overly sensitive to dropout hyperparameters and may not correctly reflect changes in data density. In contrast, GP variance predictive variance provides a probabilistic measure of uncertainty that naturally decreases in regions with high training data density. This occurs because the GP employs a kernel function to quantify the similarity between a new test input and the training examples. Fisher Information metrics capture the local sensitivity of the model's predictions to small parameter perturbations, indicating how well-determined the predictions are. An overview of this unified framework, including the flow of data from input image to the final accept/reject decision, is illustrated in Figure \ref{fig:pipeline}. By employing Particle Swarm Optimisation (PSO) to optimally combine these diverse metrics, our framework leverages their complementary strengths to yield a well-calibrated uncertainty measure. This integrated scalar not only enhances the detector's ability to reject or flag uncertain predictions, particularly when encountering novel or evolving AI-generated images, but also improves generalisation by incorporating both global data-driven confidence and local model sensitivity into a single metric. 

The key contributions of this work are as follows:

\begin{itemize}
    \item This work proposes a comprehensive framework that combines multiple uncertainty measures, namely, per instance, Fisher Information, entropy measures from MC dropout, and GP predictive variance via Deep Kernel Learning into a single, cohesive scalar uncertainty score. This integration is achieved through Particle Swarm Optimisation, which leverages the complementary strengths of each metric.
    
    \item Second, combining these uncertainty measures, our framework provides a more robust mechanism for filtering out-of-distribution (OOD) AI-generated images. This is particularly important in dynamic environments where the rapid emergence of new generative techniques may lead to images that deviate from the training distribution.
    
    \item Third, the integrated uncertainty metric not only aids in detecting novel, potentially harmful AI images but also enhances overall classifier generalisation by incorporating both global data-driven confidence (via GP variance) and local model sensitivity (via Fisher Information and MC dropout entropy).

    \item Fourth, this work demonstrates through extensive experiments that our integrated approach significantly improves the detection of OOD samples. The framework’s ability to combine multiple non-conformity measures into a single uncertainty value leads to more accurate and well-calibrated predictions compared to traditional classifiers.
\end{itemize}

\section*{Related works}

The detection of AI-generated images has gained significant attention due to the rapid advancement of generative models like GANs, diffusion models, and others. These models, including BigGAN, GLIDE, Stable Diffusion, and MidJourney, have revolutionised image synthesis but also pose risks, such as misinformation and digital forgery. Researchers have proposed diverse techniques to address the challenge of distinguishing AI-generated images from natural ones.

A notable direction has been using Convolutional Neural Networks (CNNs) for image classification, leveraging hierarchical feature representations to detect subtle differences. For instance, Zhu et al. proposed GenImage, a benchmark dataset to test the performance of classifiers in detecting synthetic images \cite{zhu2024genimage}. However, CNNs often rely on confidence scores derived from softmax outputs, which may not adequately capture out-of-distribution anomalies. Among CNN-based detection models, ResNet50  has been widely adopted to classify real and AI-generated images based on learned features\textsuperscript{\cite{wang2020cnn},\cite{gragnaniello2021gan}, \cite{ju2022fusing}}. Furthermore, the Patch Selection Module (PSM) enhances detection performance by leveraging global and local image features, integrating attention-based fusion mechanisms into the Resnet50 backbone network\cite{ju2022fusing}.
Alternative approaches employ handcrafted or statistical features. For example, spectral analysis has been explored to identify artefacts introduced during synthesis, as demonstrated by Durall et al., who analysed frequency domain differences in AI-generated images \cite{durall}. Another spectral-based approach, Deep Image Fingerprint (DIF) \cite{sinitsa2024deep}, extracts high-frequency artefacts from images using a U-Net-based high-pass filter, achieving strong generalisation even on unseen AI-generated images. Techniques like Wasserstein distances and Sinkhorn approximations have also been employed to quantify dissimilarities in feature spaces, improving detection robustness \cite{bradshaw}. Dogoulis et al. focus on the quality of the fake dataset based on the rationale that if a network is trained on high-perceptual-quality synthetic images, it will focus less on noticeable artefacts from the generative process and more on invariant characteristics across different concepts. As a result, this approach enhances the detector's generalisation ability, allowing it to distinguish AI-generated images from real ones more effectively. The study employs a ResNet-50 model for the downstream classification task of identifying real and fake images\cite{dogoulis2023improving}. However, it has been shown in previous work that they do not generalise well and failed to detect AI-generated images when the AI image generation method changes \textsuperscript{\cite{zhu2024genimage},\cite{cozzolino2018forensictransfer},\cite{zhang2019detecting},\cite{park2024performance}}. To address this challenge, online training strategies have been explored, in which detectors incrementally update their knowledge by incorporating AI-generated images in chronological order based on their historical release dates. This approach has been shown to improve performance on unseen generative models as the training history expands \cite{epstein2023online}. However, a key limitation of incremental updates is the gap between model updates and the emergence of new generative techniques. Before the detector is trained with new training data, AI-generated images produced by recently developed models may still evade detection; therefore, there is a need for uncertainty-based methods that can reject predictions when encountering data that has not yet been seen, ensuring more reliable detection in rapidly evolving AI-generation landscapes.

The application of uncertainty measures in classification is quite common. Among them, the MC dropout \cite{gal2016dropout} has been used widely to estimate uncertainty in image classification and segmentation by using dropout during inference to obtain the predictive distribution using the same input and multiple inference \textsuperscript{\cite{wang2019aleatoric}, \cite{filos2019systematic}, \cite{gour2022uncertainty}}. Although less commonly used, Fisher Information provides an alternative approach for uncertainty estimation by measuring the sensitivity of a model's predictions to small perturbations in its parameters. In this context, Fisher Information-Based Evidential Deep Learning (I-EDL) \cite{deng2023uncertainty} incorporates Fisher Information into an evidential deep learning framework to reweight the learning process based on the informativeness of predictions. By dynamically adjusting uncertainty penalties using the Fisher Information Matrix (FIM), this approach ensures that highly uncertain samples are handled more effectively. Another approach to uncertainty quantification is Deep Gaussian Processes (DGPs), which extend traditional Gaussian Processes (GPs) by stacking multiple layers of latent functions, providing a hierarchical representation of uncertainty. Recent studies have applied DGPs in radiation image reconstruction, where a Gaussian Process prior (GPP) enables Bayesian inference with structured priors, improving both image quality and uncertainty quantification \cite{lee2024radiation}. Similarly, Convolutional Deep Gaussian Processes (CDGPs) have been explored for decision-making under uncertainty in critical applications such as autonomous driving and healthcare. Furthermore, Bayesian formulations of Deep Convolutional Gaussian Processes (DCGPs) improve uncertainty calibration and model selection, offering an alternative to dropout-based Bayesian deep learning methods \cite{dutordoir2020bayesian}.

More recently, Gozalo-Salellas et al. \cite{nie2024detecting} introduced a training-free uncertainty-based framework, WePe, that detects AI-generated images by applying random weight perturbations to large pre-trained vision models such as DINOv2. Their method builds on the observation that natural and generated images exhibit different sensitivities to parameter perturbations: while the features of natural images remain relatively stable, those of generated images fluctuate more substantially, leading to higher predictive uncertainty. This approach is conceptually related to our work in that it also employs uncertainty as a detection signal. However, WePe is designed as a training-free method and directly assigns class labels based on feature instability, without incorporating a mechanism for selective rejection. In contrast, our work combines the entropy estimates derived from MC dropout, the predictive variance obtained from Gaussian Process models, and several Fisher Information-based metrics, including the total Fisher information, the Frobenius norm and Fisher entropy of the Fisher Information Matrix, to produce a single scalar uncertainty value for each sample. This composite uncertainty measure not only encapsulates the strengths of each approach but also mitigates their isolated limitations, providing a more robust assessment of model confidence.

\section*{Methodology}

\subsection*{Data Preparation and Preprocessing}
To evaluate the effectiveness of our proposed framework for detecting AI-generated images, we utilised the GenImage dataset \cite{zhu2024genimage}, a large-scale benchmark specifically designed for this purpose. GenImage comprises over 2.6 million images, evenly split between real and AI-generated content. The real images in GenImage are sourced from the ImageNet dataset, encompassing 1,331,167 images across 1,000 distinct classes. The AI-generated images are produced using eight state-of-the-art generative models, namely BigGAN \cite{brock2018large}
GLIDE \cite{nichol2021glide}, Vector Quantized Diffusion Model (VQDM) \cite{gu2022vector}
Stable Diffusion V1.4 \cite{rombach2022high}, Stable Diffusion V1.5 \cite{rombach2022high},ADM \cite{dhariwal2021diffusion}, and Midjourney \cite{midjourney2022}. Each generative model contributes approximately the same number of images per class, maintaining balance across the dataset. Typically, each model generates around 162 images per class for training and 6 for testing. An exception is Stable Diffusion V1.5, which provides slightly more, with 166 images for training and 8 for testing per class. The dataset is structured into subsets corresponding to each generative model. Each subset contains both the AI-generated images from that specific model and their corresponding real images from ImageNet for the same classes. Importantly, the real images are not shared between subsets, preventing any overlap that could introduce bias during training or evaluation. In our research, we concentrated on a specific subset of the GenImage dataset composed of images generated by Stable Diffusion V1.5, Midjourney, GLIDE, VQDM and BigGAN. To evaluate our model's performance, we exclusively trained it on images produced by Stable Diffusion V1.5 and then assessed its in-domain performance using additional images from the same generator. Meanwhile, images generated by Midjourney, GLIDE, VQDM and BigGAN were employed to gauge the model's ability to handle out-of-domain scenarios. In addition, we also included images generated by StyleGAN3. Since StyleGAN3 is trained exclusively for face generation, its distribution differs markedly from the ImageNet-based subsets in GenImage. We therefore used it solely as a supplementary, highly out-of-distribution test case to further evaluate the robustness of our framework. The StyleGAN3 subset\cite{Jha2022FakeFaceGANs} comprised 10,392 images for training and 2,598 images for testing, complemented by real face images obtained from the Deepfake Face Images dataset\cite{Bhargava_DeepfakeFaceImages_Real_2022}.

\subsection*{Statistical Analysis on Datasets}

We performed a rigorous and scientifically grounded statistical analysis to quantify the textural differences between natural (real) and AI-generated (fake) images across five datasets. Each dataset corresponds to one of five prominent generative models: \textbf{BigGAN}, \textbf{VQDM}, \textbf{Glide}, \textbf{Midjourney}, and \textbf{Stable Diffusion}. From each model, we collected 1,000 real and 1,000 AI-generated image samples, ensuring a balanced and controlled comparative setup between the two classes. This setup allowed us to make fair and interpretable comparisons between real and synthetic textures.

\begin{table}[h!]
\centering
\caption{Summary of Statistical Analysis of Textural Features}
\label{tab:stats_summary_stacked}

\vspace{1em}
\textbf{(a) Univariate Analysis (Welch's t-statistic)} \\
\vspace{0.3em}
\begin{tabular}{@{}lcccc@{}}
\toprule
\textbf{Dataset} & \textbf{Contrast} & \textbf{Energy} & \textbf{Entropy} & \textbf{Homogeneity} \\
\midrule
VQDM            & 24.50  & -2.82 & 20.02           & -19.57 \\
Midjourney      & -19.19 & -8.31 & 0.32\textsuperscript{*} & -5.40 \\
BigGAN          & 26.20  & -6.24 & 21.95           & -20.19 \\
StableDiffusion & -9.58  & 6.60  & -7.91           & 5.07 \\
Glide           & -9.58  & 6.60  & -7.91           & 5.07 \\
\bottomrule
\end{tabular}

\vspace{1em}
\textbf{(b) Multivariate Analysis (Hotelling's T\textsuperscript{2})} \\
\vspace{0.3em}
\begin{tabular}{@{}lc@{}}
\toprule
\textbf{Dataset} & \textbf{Hotelling's T\textsuperscript{2}} \\
\midrule
VQDM            & 795.16 \\
Midjourney      & 661.19 \\
BigGAN          & 849.95 \\
StableDiffusion & 155.04 \\
Glide           & 155.04 \\
\bottomrule
\end{tabular}

\vspace{1em}
\textbf{(c) Distributional Drift (KL Divergence)} \\
\vspace{0.3em}
\begin{tabular}{@{}lcccc@{}}
\toprule
\textbf{Dataset} & \textbf{Contrast} & \textbf{Energy} & \textbf{Entropy} & \textbf{Homogeneity} \\
\midrule
VQDM            & 0.01 & 39.57 & 2.54 & 36.94 \\
Midjourney      & 0.00 & 12.68 & 0.83 & 7.48 \\
BigGAN          & 0.02 & 36.69 & 6.95 & 33.67 \\
StableDiffusion & 0.00 & 58.15 & 1.22 & 43.47 \\
Glide           & 0.00 & 58.15 & 1.22 & 43.47 \\
\bottomrule
\end{tabular}

\vspace{0.8em}
\begin{flushleft}
\footnotesize
\textsuperscript{*} $p = 0.7477$, indicating no significant difference. For all other t-tests and all Hotelling's T\textsuperscript{2} tests, $p < 0.005$.
\end{flushleft}
\end{table}

\begin{figure}[h]
    \centering
    \includegraphics[width=\linewidth]{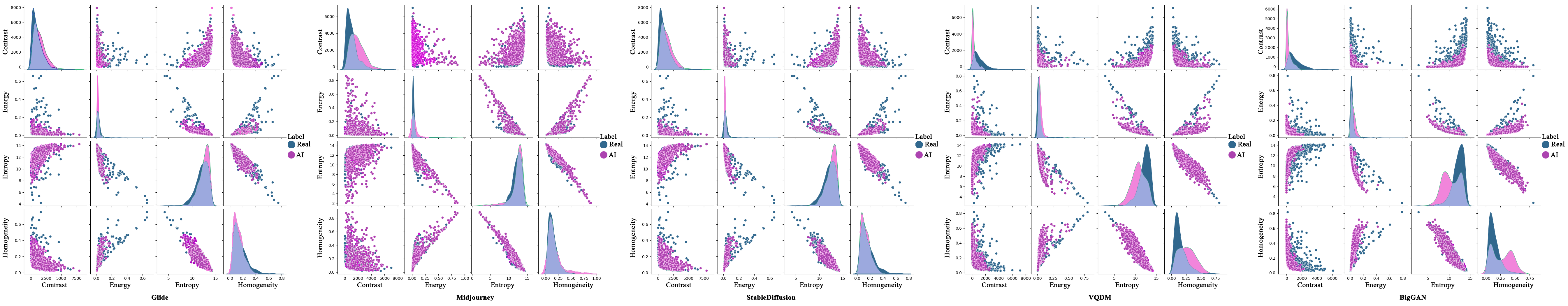}
    \caption{Pairplots comparing real (blue) and AI-generated (purple) images across four GLCM features: Contrast, Energy, Entropy, and Homogeneity. Each subplot shows the feature distribution and inter-feature relationships for a specific model.}
    \label{fig:combine_pairplots}
\end{figure}

Our methodology is based on the hypothesis that generative models, despite their impressive visual quality, introduce subtle yet consistent textural artefacts that deviate from the statistical properties of real images. To investigate this, we employed the Grey-Level Co-occurrence Matrix (GLCM) method, which quantifies spatial relationships between pixel intensities and provides interpretable measures of image texture. From each image, we extracted four key features: \textit{Contrast}, \textit{Energy}, \textit{Entropy}, and \textit{Homogeneity}. These features serve as a compact representation of each image’s texture signature.

To statistically evaluate whether these features differ between real and AI-generated images, we applied three complementary analytical techniques. First, we independently conducted a univariate analysis using Welch’s t-test for each feature. This test is particularly appropriate because it does not assume equal variances between groups. The results, summarised in Table~\ref{tab:stats_summary_stacked}, revealed strong evidence of significant differences in most features across datasets. For instance, in the VQDM dataset, Welch’s t-test showed a t-statistic of 24.50 for contrast and 20.02 for entropy, both with $p < 0.0001$, indicating highly significant differences. Interestingly, the entropy values in the Midjourney dataset were not significantly different between real and AI images ($t = 0.32$, $p = 0.7477$), suggesting that this specific generative model produces images with entropy levels more closely aligned to natural images.

Second, we conducted a multivariate analysis using Hotelling’s T\textsuperscript{2} test to assess the joint significance of all four GLCM features. This multivariate test evaluates whether the combined distribution of features for real images differs from that of AI-generated images. As shown in Table~\ref{tab:stats_summary_stacked}, all five models yielded highly significant results ($p < 0.0001$), with particularly high T\textsuperscript{2} statistics for BigGAN (849.95) and VQDM (795.16), indicating substantial multivariate separation between real and synthetic images.

Third, to measure the degree of distributional drift beyond hypothesis testing, we calculated the Kullback-Leibler (KL) divergence between the feature distributions of real and AI-generated images. This metric quantifies how much the probability distribution of each feature in AI-generated images deviates from its real counterpart. Notably, the energy feature showed large KL divergence in multiple models—for example, Stable Diffusion and Glide both had Energy KL values of 58.15—highlighting significant shifts in the underlying textural distributions, as reported in Table~\ref{tab:stats_summary_stacked}.

These statistical findings are further supported by the pairplot visualisations presented in Figure~\ref{fig:combine_pairplots}, which illustrate the inter-feature relationships and distributional differences between the real and AI-generated classes for each dataset. In these plots, real images are shown in blue and AI-generated images in green. The degree of clustering or overlap visually reinforces the results observed in the statistical tests. Glide, for instance, demonstrates moderate overlap in entropy and homogeneity but shows clear distinctions in contrast and Energy. Midjourney displays greater separation in contrast and Energy but minimal difference in entropy, aligning with the non-significant p-value observed. Stable Diffusion exhibits some overlapping, but shows lower contrast and Energy in AI-generated samples. In contrast, VQDM and BigGAN both show clear separation across all four features, indicating highly distinguishable synthetic textures.

Overall, this combination of statistical analysis and visual evidence provides robust support for the claim that AI-generated images exhibit distinct and quantifiable differences from natural images in terms of their textural properties.

\subsection*{Base Classifier}
Our framework leverages existing base classifiers, fine-tuning them to differentiate between AI-generated and natural images. Our primary model is a trained ResNet50 architecture \textsuperscript{\cite{he2016deep}}, which has been extensively employed in AI versus natural image detection \textsuperscript{\cite{wang2020cnn},\cite{gragnaniello2021gan},\cite{ju2022fusing}}. By capitalising on the hierarchical features learned from large-scale datasets like ImageNet, we retain the convolutional base of ResNet50 as a robust feature extractor. The original classification head is replaced with a custom-designed module tailored for our binary classification task. This module begins with a global average pooling layer that condenses the feature maps into a 2048-dimensional vector, which is then processed through a classification head consisting of two fully connected layers with 512 and 256 units, respectively. Each fully connected layer is accompanied by ReLU activations and dropout regularisation (with dropout rates of 0.5) to mitigate overfitting, culminating in a final layer that outputs a single logit for decision-making.

In addition to our ResNet50-based approach, we also explored a Vision Transformer (ViT) model to assess the generalizability of our uncertainty measures across diverse architectures. For the ViT model, we employed the vit\_base\_patch16\_224 architecture, renowned for its effective feature extraction. Similar to the ResNet50 adaptation, the original classification head of the ViT was removed and substituted with a custom-designed classifier. The ViT extracts a 768-dimensional embedding from its feature extraction stage, which is then reduced to 512 units via a fully connected layer, followed by ReLU activation and dropout. This intermediate representation is further refined by another fully connected layer, reducing it to 256 units, again using ReLU activation and dropout, before a final fully connected layer outputs a single logit for binary classification. In order to maintain the integrity of the pre-trained representations, most of the ViT layers are kept frozen during training, with only the final two transformer blocks being fine-tuned to capture task-specific nuances.

By integrating both the ResNet50 and ViT architectures within our framework, we demonstrate that our approach is robust and adaptable to different neural architectures. While our primary focus is on the extensively validated ResNet50, the experiments with ViT show the potential of our uncertainty measures to enhance classification performance across various models.

\subsection*{Integration of Uncertainity Measure}
\subsubsection*{Per Instance Fisher Information}

Fisher Information serves as the first uncertainty measure in our framework, quantifying how each image, whether AI-generated or natural, influences the parameters of our classification model. This metric assesses the sensitivity of the model's likelihood function to changes in its parameters on a per-instance basis, thereby providing insights into how well an image conforms to the learned patterns.

The foundation of this approach is the Fisher Information Matrix (FIM) for a single data point, which quantifies the amount of information the observable data carries about the model parameters. For a given data point \((x_i, y_i)\) and model parameters \(\theta\), the per-instance Fisher Information is defined as:

\begin{equation}
    I_i(\theta) = \mathbb{E}_{y}\left[ \left( \nabla_\theta \log p(y | x_i; \theta) \right) \left( \nabla_\theta \log p(y | x_i; \theta) \right)^\top \right],
\end{equation}

where \(p(y | x_i; \theta)\) represents the likelihood of observing the label \(y\) given the input \(x_i\) under parameters \(\theta\), and \(\nabla_\theta\) denotes the gradient with respect to \(\theta\).

Due to the high dimensionality of deep neural network parameters, computing the full Fisher Information Matrix (FIM) for every layer is highly resource-intensive. To simplify this, we concentrate on the customs classification head that sits atop our pre-trained ResNet50 and ViT models. This part of the network is solely responsible for making the final binary decision between AI-generated and natural images. By extracting the FIM from just this custom head, we can capture the most important information about how individual inputs influence the model's decisions without overwhelming our computational resources.

For each input image \(x_i\), we perform a forward pass to obtain the output logits \(f(x_i; \theta)\). We then compute the gradients of the binary cross-entropy loss \(\mathcal{L}(y, f(x_i; \theta))\) with respect to the model parameters for each potential class label \(y \in \{0, 1\}\), regardless of the true label. We assume equal prior probabilities for both classes, \(p(0) = p(1) = 0.5\), and the gradient for each class is given by:

\begin{equation}
g_i^{(y)} = \nabla_\theta \mathcal{L}(y, f(x_i; \theta)).
\end{equation}

The per-instance Fisher Information Matrix is then approximated as:

\begin{equation}
I_i(\theta) = \sum_{y \in \{0, 1\}} p(y) \left( g_i^{(y)} \odot g_i^{(y)} \right),
\end{equation}

where \(\odot\) denotes element-wise multiplication. This formulation effectively accumulates the squared gradients weighted by the class probabilities, capturing each parameter's expected curvature of the loss function.

To distil this matrix into a set of scalar metrics that summarise the uncertainty associated with each instance, we extract several key measures:
\begin{itemize}
    \item Total Fisher Information (Trace):  This is computed as
  \begin{equation}
  \text{Trace}(I_i(\theta)) = \sum_{k} \left[ I_i(\theta) \right]_k,
  \end{equation}
  where \(\left[ I_i(\theta) \right]_k\) represents the diagonal elements of the matrix, corresponding to the variance of each parameter.
  \item  Frobenius Norm:  This metric captures the overall magnitude of the Fisher Information Matrix, calculated as
  \begin{equation}
  \| I_i(\theta) \|_F = \sqrt{ \sum_{k} \left( \left[ I_i(\theta) \right]_k \right)^2 }.
  \end{equation}
  \item Entropy-Based Measure:  By normalizing the diagonal elements, we define a probability distribution \(p_i(k) = \frac{ \left[ I_i(\theta) \right]_k }{ \text{Trace}(I_i(\theta)) }\) and compute the entropy as:
  \begin{equation}
  \text{Entropy} = - \sum_{k} p_i(k) \log \left( p_i(k) + \epsilon \right),
  \end{equation}
  where \(\epsilon\) is a small constant for numerical stability.
\end{itemize}

By concentrating on the parameters of the custom classification head, we directly assess the sensitivity of the model's decision-making process. In this architecture, the pre-trained convolutional base extracts general features from the images, and the classification head interprets these features to produce task-specific predictions. Consequently, the per-instance Fisher Information derived from the custom head clearly measures how each image affects the model's capacity to distinguish between AI-generated and natural images. Lower Fisher Information values suggest that an image exerts minimal influence on the model parameters, potentially indicating non-conformity with learned patterns. In contrast, higher values indicate that the image strongly aligns with the training data. Finally, these scalar metrics are aggregated across the relevant layers of the classification head. We sum the values from all selected layers for cumulative metrics, such as the Total Fisher Information and Frobenius Norm. In contrast, we compute the mean across layers for metrics expressed as averages, such as the entropy-based measure. 

\subsubsection*{Deep Kernel Learning with Gaussian Processes}

Deep Kernel Learning (DKL) provides a robust hybrid framework that combines the representational strength of deep neural networks with the probabilistic flexibility of Gaussian Processes (GPs). This approach enables the extraction of rich, semantically meaningful features from high-dimensional image data while leveraging GPs to model uncertainty and make robust predictions. In our work, DKL is employed to classify images into two categories: AI-generated and natural images, with the variance of the GP posterior serving as our primary measure of uncertainty.

Due to the complexity and high dimensionality of image data, we first transform raw pixel inputs into a compact and semantically meaningful feature space using a deep neural network. This transformation is achieved through a modified ResNet50 or ViT network trained as a feature extractor. Let the neural network feature extractor be denoted as  
 \begin{equation}
\mathcal{F}_\theta: \mathbb{R}^{\text{image dims}} \to \mathbb{R}^d,
 \end{equation}  
parameterized by \(\theta\). For an image \(\mathbf{x}\), the extracted feature vector is:
 \begin{equation}
\mathbf{z} = \mathcal{F}_\theta(\mathbf{x}),
 \end{equation}
where \(\mathbf{z} \in \mathbb{R}^d\) represents the embedding of the learned feature. These embeddings are designed to capture the essential distinctions between AI-generated and natural images, reducing the complexity of the input space while preserving discriminative information. To model the classification task, we place a Gaussian Process prior to a latent function \( f: \mathbb{R}^d \to \mathbb{R} \) that maps the feature vectors \(\mathbf{z}\) to a real-valued latent score. The GP prior is defined as:
 \begin{equation}
f(\mathbf{z}) \sim \mathcal{GP}(m(\mathbf{z}), k(\mathbf{z}, \mathbf{z}')),
 \end{equation}
where \(m(\mathbf{z})\) is the mean function, commonly assumed to be zero, and \(k(\mathbf{z}, \mathbf{z}')\) is a kernel function that encodes similarity between feature vectors. By setting \(m(\mathbf{z}) = 0\), the prior simplifies to:
 \begin{equation}
f(\mathbf{z}) \sim \mathcal{GP}(0, k(\mathbf{z}, \mathbf{z}')).
 \end{equation}

In our implementation, we use the Radial Basis Function (RBF) kernel given by:
 \begin{equation}
k(\mathbf{z}, \mathbf{z}') = \sigma^2 \exp\left( -\frac{\|\mathbf{z} - \mathbf{z}'\|^2}{2l^2}\right),
 \end{equation}
Where \(\sigma^2\) is the output scale and \(l\) is the length scale, both of which are learned during training. By combining the deep feature extractor with the RBF kernel, we effectively learn a deep kernel that adapts to the specific characteristics of our data.

Although GPs provide a flexible approach to modelling functions, their computational cost scales cubically with the number of training points \(N\), making the exact GP inference infeasible for large datasets. To address this limitation, we introduce \(M \ll N\) inducing points. 
 \begin{equation}
\mathbf{Z}_u = [\mathbf{z}_u^{(1)}, \ldots, \mathbf{z}_u^{(M)}],
 \end{equation}
which acts as a low-rank approximation of the GP. These inducing points summarise the function's behaviour over the input space, reducing computational complexity while retaining much of the GP’s expressive power. In our implementation, we use 1000 inducing points, 500 for each class. The corresponding inducing function values 
 \begin{equation}
\mathbf{u} = [f(\mathbf{z}_u^{(1)}), \ldots, f(\mathbf{z}_u^{(M)})]^\top
 \end{equation}
are assumed to follow the same GP prior:
 \begin{equation}
\mathbf{u} \sim \mathcal{N}(\mathbf{0}, \mathbf{K}_{u,u}),
 \end{equation}
where \(\mathbf{K}_{u,u}\) is the kernel matrix evaluated on the inducing points. The full latent function values \(\mathbf{f}\) for the training data are then conditionally Gaussian given \(\mathbf{u}\):
 \begin{equation}
\mathbf{f}|\mathbf{u} \sim \mathcal{N}(\mathbf{K}_{f,u}\mathbf{K}_{u,u}^{-1}\mathbf{u}, \mathbf{K}_{f,f} - \mathbf{K}_{f,u}\mathbf{K}_{u,u}^{-1}\mathbf{K}_{u,f}),
 \end{equation}
Where \(\mathbf{K}_{f,u}\) and \(\mathbf{K}_{f,f}\) are kernel matrices evaluated between training points and inducing points, and among training points alone, respectively. This sparse approximation reduces the cost of GP inference and training to \(\mathcal{O}(NM^2)\), making it feasible for larger datasets.

Since this is a binary classification problem, we map the latent function values \(f(\mathbf{z})\) to class probabilities using the logistic sigmoid function:
 \begin{equation}
\sigma(f) = \frac{1}{1 + \exp(-f)}.
 \end{equation}
The likelihood of a label \(y \in \{0,1\}\) given \(f(\mathbf{z})\) is:
 \begin{equation}
p(y|f) = \sigma(f)^{y}[1-\sigma(f)]^{1-y}.
 \end{equation}
For a dataset with \(N\) training points, the joint likelihood is:
 \begin{equation}
p(\mathbf{y}|\mathbf{f}) = \prod_{i=1}^N p(y_i|f_i).
 \end{equation}

Because this likelihood is non-Gaussian, exact posterior inference is intractable. Instead, we employ variational inference, where we approximate the posterior \(p(\mathbf{f}, \mathbf{u}|\mathbf{y})\) using a variational distribution:
 \begin{equation}
q(\mathbf{f}, \mathbf{u}) = p(\mathbf{f}|\mathbf{u})q(\mathbf{u}),
 \end{equation}
and \(q(\mathbf{u})\) is chosen to be Gaussian:
 \begin{equation}
q(\mathbf{u}) = \mathcal{N}(\mathbf{u}|\boldsymbol{\mu}, \mathbf{S}).
 \end{equation}

The Evidence Lower Bound (ELBO) provides a tractable objective for optimising the variational parameters \(\boldsymbol{\mu}\) and \(\mathbf{S}\):
 \begin{equation}
\mathcal{L} = \mathbb{E}_{q(\mathbf{f})}[\log p(\mathbf{y}|\mathbf{f})] - \mathrm{KL}[q(\mathbf{u})||p(\mathbf{u})].
 \end{equation}
Here, the first term measures how well the variational approximation explains the data. In contrast, the second term regularises the variational posterior \(q(\mathbf{u})\), ensuring it remains consistent with the GP prior. By maximising \(\mathcal{L}\), we jointly optimise the kernel hyperparameters, the inducing point locations, and the deep feature extractor parameters \(\theta\).
For a test point \(\mathbf{x}_*\), we compute the predictive distribution over the latent function \(f(\mathbf{z}_*)\) using the GP posterior. The resulting distribution is Gaussian:
 \begin{equation}
q(f_*) = \mathcal{N}(f_*|\mu_*, \sigma_*^2),
 \end{equation}
where \(\mu_*\) and \(\sigma_*^2\) are derived from the variational posterior. Importantly, the variance \(\sigma_*^2\) serves as our uncertainty measure, quantifying the model's confidence in its prediction. The class probabilities are obtained by integrating over this predictive distribution:
 \begin{equation}
p(y_* = 1|\mathbf{z}_*) = \int \sigma(f_*)q(f_*) \, df_*.
 \end{equation}

Due to the nonlinearity of the sigmoid function, this integral lacks a closed-form solution. To approximate it, we employ Monte Carlo methods. In this approach, we draw \(S\) latent function samples \(f_*^{(s)}\) from the Gaussian distribution \(q(f_*) = \mathcal{N}(f_*|\mu_*, \sigma_*^2)\), where each sample is generated as:
 \begin{equation}
f_*^{(s)} = \mu_* + \epsilon^{(s)} \sigma_*,
 \end{equation}
and \(\epsilon^{(s)}\) is drawn from a standard normal distribution \(\mathcal{N}(0, 1)\). For each sample, we evaluate the sigmoid function to produce probabilities \(\sigma(f_*^{(s)})\). The integral is then approximated by averaging these probabilities:
 \begin{equation}
p(y_* = 1|\mathbf{z}_*) \approx \frac{1}{S} \sum_{s=1}^S \sigma(f_*^{(s)}).
 \end{equation}

This Monte Carlo approximation not only provides the predicted class probabilities but also captures predictive uncertainty through the variance \(\sigma_*^2\) of the GP posterior. This variance is used as uncertainty in the framework, as it informs us about the confidence of the model's predictions and enables robust decision-making in our classification task.

\subsubsection*{MC Dropout}

MC Dropout is integrated into our framework to quantify uncertainty in both the ResNet50-based and ViT-based classifiers. MC Dropout is applied during inference to generate a distribution of probabilistic outputs through multiple stochastic forward passes. Dropout layers within the custom classification head of Resnet50 and ViT remain activated during inference, enabling the model to produce \(N = 20\) stochastic predictions for each input image. The multiple outputs obtained via MC Dropout are aggregated to compute key uncertainty metrics. The Entropy of Expected is calculated as:

\begin{equation}
\text{Entropy of Expected} = -\bar{p} \log(\bar{p}) - (1-\bar{p}) \log(1-\bar{p}),
 \end{equation}

Where \(\bar{p}\) is the mean predicted probability across all MC passes. We also compute the Expected Entropy:

\begin{equation}
\text{Expected Entropy} = -\frac{1}{N} \sum_{i=1}^{N} \left( p_i \log(p_i) + (1 - p_i) \log(1 - p_i) \right),
 \end{equation}

with \(p_i\) representing the predicted probability from the \(i\)th forward pass. The difference between these two metrics, defined as Knowledge Uncertainty,

\begin{equation}
\text{Knowledge Uncertainty} = \text{Entropy of Expected} - \text{Expected Entropy},
 \end{equation}

The rationale for using MC Dropout lies in its ability to capture the variability in model predictions that arises from uncertainty. For instance, the Entropy of Expected metric reflects the uncertainty in the model's parameters, typically peaking in regions with sparse or unseen data. In contrast, Expected Entropy highlights the inherent noise within the data, such as overlapping classes or noisy labels. The difference between these two metrics, termed knowledge uncertainty, offers additional insight by identifying instances where uncertainty arises from a combination of data ambiguity and model limitations. In our experiments, we observed that Expected Entropy yields nearly the same values as the Entropy of Expected; therefore, in the overall framework, we use the Entropy of Expected and Knowledge uncertainty as the uncertainty derived from MC dropout in the overall framework.

\subsection*{Combining and Normalising Uncertainty Measures}

In this section, we describe how to combine the uncertainty metrics introduced earlier (Fisher-based measures, Gaussian Process variance, and MC Dropout-based metrics) into a single unified uncertainty score per sample. The goal is to leverage a weighted combination of these measures so that samples with high overall uncertainty can be rejected rather than incorrectly accepted. To optimise the weights of each uncertainty measure and the rejection threshold, we employ Particle Swarm Optimisation (PSO).

From the previous section, recall that each sample \(i\) is associated with multiple uncertainty measures: Fisher-based uncertainties, Gaussian Process (GP) variance, and MC Dropout metrics. To ensure higher uncertainty values reflect greater non-conformity, the Fisher-based measures are transformed by taking their reciprocals:
\begin{itemize}
    \item Fisher Total Uncertainty:
  \begin{equation}
      \text{Fisher Total Uncertainty} = \frac{1}{\text{Trace}(I_i(\theta)) + \epsilon},
   \end{equation}
  \item Fisher Frobenius Uncertainty:
  \begin{equation}
      \text{Fisher Frobenius Uncertainty} = \frac{1}{\| I_i(\theta) \|_F + \epsilon},
   \end{equation}

    \item Fisher Entropy Uncertainty:
  \begin{equation}
      \text{Fisher Entropy Uncertainty} = \frac{1}{\text{Fisher Entropy} + \epsilon}.
   \end{equation}
\end{itemize}

The  GP variance is directly used as an uncertainty measure. Finally,  MC Dropout provides two additional metrics (e.g., Entropy of Expected Predictions and Knowledge Uncertainty). All these measures 
are column-stacked into a single feature matrix \(\mathbf{X}_{\text{uncertainty}}\). Z-score normalisation is applied to each feature to balance the contributions of each uncertainty measure.

 \begin{equation}
    x_{\text{norm}} = \frac{x - \mu}{\sigma},
 \end{equation}

yielding the normalised matrix

 \begin{equation}
\mathbf{X}_{\text{norm}} =
\begin{bmatrix}
x_{1,1}^{\text{norm}} & x_{1,2}^{\text{norm}} & \ldots & x_{1,6}^{\text{norm}} \\
x_{2,1}^{\text{norm}} & x_{2,2}^{\text{norm}} & \ldots & x_{2,6}^{\text{norm}} \\
\vdots & \vdots & \ddots & \vdots \\
x_{N,1}^{\text{norm}} & x_{N,2}^{\text{norm}} & \ldots & x_{N,6}^{\text{norm}}
\end{bmatrix}.
 \end{equation}

Here, \(x\) is the raw uncertainty value, \(\mu\) is its mean, and \(\sigma\) is the standard deviation for that feature. A linear combination of the normalized features, weighted by \(\mathbf{w}\), produces a single combined uncertainty score \(u_i\) for each sample:

 \begin{equation}
    u_i = \sum_{j=1}^m w_j \cdot x_{i,j},
 \end{equation}

Which can be written in matrix form as:

 \begin{equation}
  \mathbf{u} = \mathbf{X}_{\text{norm}} \cdot \mathbf{w}.
 \end{equation}

A sample is rejected if its combined score \(u_i\) exceeds a threshold \(\tau\):

 \begin{equation}
\text{Rejected} = \{ i \mid u_i > \tau \}.
 \end{equation}
To optimise the weights \(\mathbf{w}\) and the rejection threshold \(\tau\), we use Particle Swarm Optimisation (PSO). The objective function aims to maximise the sum of Correctly Predicted Acceptance (CPA\%) and Incorrectly Predicted Rejection (IPR\%) rates for both AI and Nature, given by the following formulae.

   \begin{equation}
      \text{CPA\%} = \frac{\text{CA}}{\text{CA} + \text{CR}},
   \end{equation}
  where \(\text{CA}\) is the number of correctly classified samples with \(u_i \leq \tau\).

   \begin{equation}
      \text{IPR\%} = \frac{\text{CR}}{\text{CR} + \text{IA}},
   \end{equation}
  where \(\text{CR}\) is the number of misclassified samples with \(u_i > \tau\), and \(\text{IA}\) the misclassified samples with \(u_i \leq \tau\).

The total score to be maximised is:

 \begin{equation}
\text{Score} = \text{CPA\%}_{\text{AI}} + \text{IPR\%}_{\text{AI}} + \text{CPA\%}_{\text{Nature}} + \text{IPR\%}_{\text{Nature}}.
 \end{equation}

Because PSO solves a minimisation problem, we minimise the negative of this score:

 \begin{equation}
(\mathbf{w}^*, \tau^*) = \arg\min_{\mathbf{w}, \tau} \Bigl(- \sum_{\text{classes}} \bigl(\text{CA\%} + \text{CR\%}\bigr)\Bigr).
 \end{equation}

Each PSO particle \(\mathbf{p}\) is defined as:

 \begin{equation}
\mathbf{p} = [w_1, w_2, \ldots, w_m, \tau],
 \end{equation}

With positions (weights and threshold) initialised randomly within predefined bounds \(\bigl[w_j^\text{min}, w_j^\text{max}\bigr]\) and \([\tau^\text{min}, \tau^\text{max}]\). At each iteration, the velocity \(\mathbf{v}_i\) and position \(\mathbf{p}_i\) of particle \(i\) are updated:

 \begin{equation}
\mathbf{v}_i = \omega \mathbf{v}_i 
    + \phi_p \mathbf{r}_p \odot (\mathbf{p}_i^{\text{best}} - \mathbf{p}_i) 
    + \phi_g \mathbf{r}_g \odot (\mathbf{g}^{\text{best}} - \mathbf{p}_i),
 \end{equation}

 \begin{equation}
\mathbf{p}_i = \mathbf{p}_i + \mathbf{v}_i.
 \end{equation}

\begin{figure}
    \centering
    \includegraphics[width=0.7\textwidth]{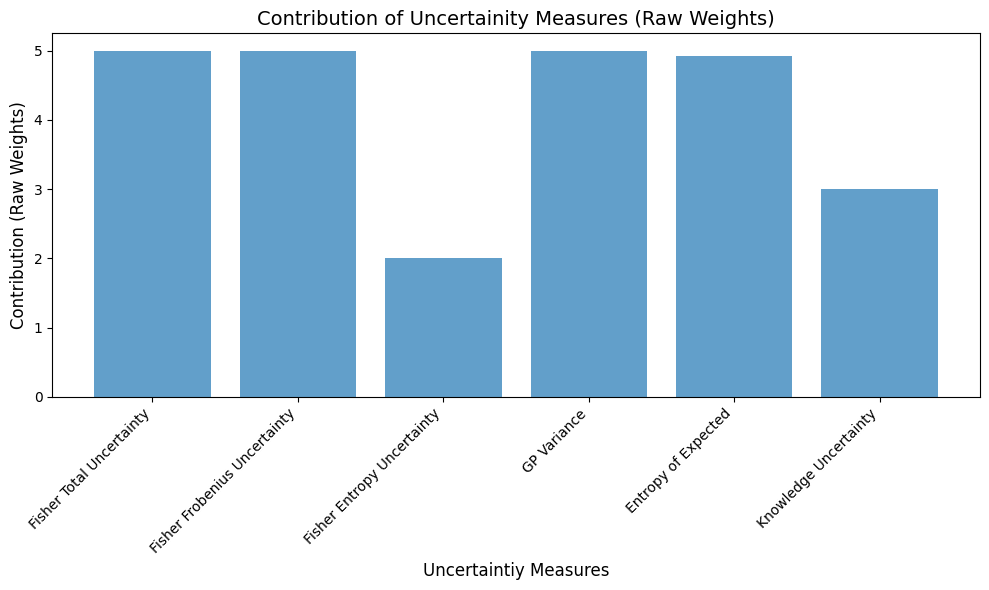}
    \caption{ PSO-derived weighting scheme for combining multiple uncertainty measures from Resnet50.}
    \label{fig:pso_weights}
\end{figure}

Here, \(\omega\) is the inertia weight, \(\phi_p\) and \(\phi_g\) are cognitive and social coefficients, \(\mathbf{r}_p\) and \(\mathbf{r}_g\) are random vectors in \([0,1]\), and \(\odot\) denotes elementwise multiplication. Positions are clipped to respect the specified bounds, and velocities are reset if they reach the boundaries. Upon convergence, the PSO returns the optimal weights \(\mathbf{w}^*\), the optimal threshold \(\tau^*\), and the highest final score. These weights determine the relative importance of each uncertainty measure, and \(\tau^*\) serves as the unified decision boundary for rejecting samples with high combined uncertainty. Figure \ref{fig:pso_weights} illustrates the final weights \(\mathbf{w}^*\) discovered by PSO for each uncertainty measure derived from Resnet50. Notably, Total Fisher Information and GP variance receive higher weights, indicating their more influential role in determining whether a sample is ultimately accepted or rejected. By contrast, MC Dropout and other measures contribute less to this setup. Once derived, these PSO-based weights can be applied consistently across all datasets, providing a unified acceptance–rejection mechanism that generalises under diverse data distributions.

\section*{Results}

We first evaluated the model using data drawn from the same source as the training set, partitioning that dataset into separate training and testing splits. Table \ref{tab:performance_metrics_same_split} reports the resulting precision, recall and F1 scores in the Resnet50 model, while Table \ref{tab:confusion_matrix_same_split} presents the corresponding confusion matrix. To examine performance on images generated by different techniques from what the model is trained on, we next tested the model trained solely on Stable Diffusion data against four additional datasets (Midjourney, Glide, VQDM and BigGAN). The results of these cross-dataset experiments are summarised in Table \ref{tab:cross_resnet_performance} for Resnet50 and Table \ref{tab:cross_vit_performance} for ViT. As shown in Table \ref{tab:cross_resnet_performance} and Table \ref{tab:conf-matrix-data-shift-vit}, the detection accuracy decreases significantly when applied to previously unseen generative models, and the confusion matrix in Tables \ref{tab:conf-matrix-data-shift-res} and \ref{tab:conf-matrix-data-shift-vit} indicates that the majority of misclassifications occur for AI-generated images.

\begin{table}
\centering
\caption{Performance Metrics for Each Dataset with test set drawn from the same AI image generation technique (AI = 0, Nature = 1) using ResNet50}
\label{tab:performance_metrics_same_split}
\begin{tabular}{|l|c|c|c|c|}
\hline
\textbf{Dataset} & \textbf{Accuracy} & \textbf{Precision} & \textbf{Recall} & \textbf{F1 Score} \\ \hline
Stable Diffusion & 0.9979 & 0.9976 & 0.9982 & 0.9979 \\ \hline
Mid Journey      & 0.9892 & 0.9938 & 0.9845 & 0.9891 \\ \hline
Glide            & 0.9936 & 0.9878 & 0.9995 & 0.9936 \\ \hline
VQDM             & 0.9988 & 0.9977 & 0.9998 & 0.9988 \\ \hline
BigGAN          & 0.9978 & 0.9970 & 0.9985 & 0.9978 \\ \hline
StyleGAN3        & 0.9978 & 0.9970 & 0.9985 & 0.9978 \\ \hline

\end{tabular}
\end{table}

%
%
\begin{table}
\centering
\caption{Confusion Matrices for Each Dataset with test set drawn from the same AI image generation technique using ResNet50 (AI=0, Nature=1). TN = AI $\to$ AI, FP = AI $\to$ Nature, FN = Nature $\to$ AI, TP = Nature $\to$ Nature}
\label{tab:confusion_matrix_same_split}
\begin{tabular}{|l|c|c|c|c|}
\hline
\textbf{Dataset} & \textbf{TN (0$\to$0)} & \textbf{FP (0$\to$1)} & \textbf{FN (1$\to$0)} & \textbf{TP (1$\to$1)} \\ \hline
Stable Diffusion & 7981 & 19  & 14  & 7986 \\ \hline
Mid Journey      & 5963 & 37  & 93  & 5907 \\ \hline
Glide            & 5926 & 74  & 3   & 5997 \\ \hline
VQDM             & 5986 & 14  & 1   & 5999 \\ \hline
BigGAN           & 5982 & 18  & 9   & 5991 \\ \hline
StyleGAN3 & 1404 & 16 & 1 & 1177 \\ \hline
\end{tabular}
\end{table}

\begin{table}
\centering
\caption{Performance Metrics for each Dataset when tested on ResNet50 model trained with Stable Diffusion}
\label{tab:cross_resnet_performance}
\begin{tabular}{|l|c|c|c|c|}
\hline
\textbf{Dataset} & \textbf{Accuracy} & \textbf{Precision} & \textbf{Recall} & \textbf{F1 Score} \\ \hline
Midjourney            & 0.6602 & 0.5957 & 0.9972 & 0.7458 \\ \hline
GLIDE                 & 0.7361 & 0.6550 & 0.9977 & 0.7908 \\ \hline
VQDM                  & 0.5717 & 0.5387 & 0.9965 & 0.6994 \\ \hline
BigGAN                & 0.5597 & 0.5319 & 0.9975 & 0.6938 \\ \hline
StyleGAN3            & 0.4534 & 0.4534 & 1.0000 & 0.6239 \\ \hline
\end{tabular}
\end{table}

\begin{table}
\centering
\caption{Performance Metrics for each Dataset when tested on ViT model trained with stable diffusion}
\label{tab:cross_vit_performance}
\begin{tabular}{|l|c|c|c|c|}
\hline
\textbf{Dataset} & \textbf{Accuracy} & \textbf{Precision} & \textbf{Recall} & \textbf{F1 Score} \\ \hline
Stable Diffusion & 0.9709 & 0.9753 & 0.9663 & 0.9707 \\ \hline
Mid Journey      & 0.6454 & 0.5885 & 0.9672 & 0.7317 \\ \hline
Glide            & 0.6148 & 0.5674 & 0.9660 & 0.7149 \\ \hline
VQDM             & 0.5479 & 0.5259 & 0.9713 & 0.6824 \\ \hline
BigGAN           & 0.5130 & 0.5068 & 0.9667 & 0.6650 \\ \hline
StyleGAN3        & 0.4831 & 0.4637 & 0.8939 & 0.6106 \\ \hline
\end{tabular}
\end{table}

\begin{table}
\centering
\caption{Confusion Matrices for each Dataset when tested on ResNet50 model trained with Stable Diffusion (AI=0, Nature=1). TN = AI $\to$ AI, FP = AI $\to$ Nature, FN = Nature $\to$ AI, TP = Nature $\to$ Nature}
\label{tab:conf-matrix-data-shift-res}
\begin{tabular}{|l|c|c|c|c|}
\hline
\textbf{Dataset} & \textbf{TN (0$\to$0)} & \textbf{FP (0$\to$1)} & \textbf{FN (1$\to$0)} & \textbf{TP (1$\to$1)} \\ \hline
Midjourney            & 1939 & 4061 & 17 & 5983 \\ \hline
GLIDE                 & 2847 & 3153 & 14 & 5986 \\ \hline
VQDM                  & 881  & 5119 & 21 & 5979 \\ \hline
BigGAN           & 732  & 5268 & 15 & 5985 \\
\hline
StyleGAN3        & 0    & 1420 & 0  & 1178 \\ \hline

\hline
\end{tabular}
\end{table}

\begin{table}
\centering
\caption{Confusion Matrices for each Dataset when tested on ViT model trained with stable diffusion (AI=0, Nature=1). TN = AI $\to$ AI, FP = AI $\to$ Nature, FN = Nature $\to$ AI, TP = Nature $\to$ Nature}
\label{tab:conf-matrix-data-shift-vit}
\begin{tabular}{|l|c|c|c|c|}
\hline
\textbf{Dataset} & \textbf{TN (0$\to$0)} & \textbf{FP (0$\to$1)} & \textbf{FN (1$\to$0)} & \textbf{TP (1$\to$1)} \\ \hline
Stable Diffusion & 7804 & 196  & 270  & 7730 \\ \hline
Mid Journey      & 1942 & 4058 & 197  & 5803 \\ \hline
Glide            & 1581 & 4419 & 204  & 5796 \\ \hline
VQDM             &  747 & 5253 & 172  & 5828 \\ \hline
BigGAN          &  356 & 5644 & 200  & 5800 \\ \hline
StyleGAN3       &  202 & 1218 & 125 & 1053 \\ \hline
\end{tabular}
\end{table}

To address the limitations inherent in standard classification, we introduced an acceptance-rejection mechanism governed by a threshold \(\tau\), where the model rejects samples if their confidence (or, conversely, uncertainty) does not meet this specified threshold. Each sample \(i\) in the dataset is assigned a scalar score \(u_i\), which can be interpreted as either uncertainty (higher values indicating lower confidence) or confidence (higher values indicating greater confidence). By accepting only those samples that exceed the confidence threshold, the system reduces the risk of erroneously labelling AI-generated images as Natural Images. The model either accepts or rejects the sample based on whether \(u_i\) crosses a threshold \(\tau\). Formally, one may reject a sample if \(u_i > \tau\) (for uncertainty-based scoring) or if \(u_i < \tau\) (for confidence-based scoring). We define four primary outcomes, and they are Correct Accepted (CA) which is the number of samples for which the prediction is correct and the sample is accepted by the threshold rule, Correct Rejected (CR), which is the number of samples for which the prediction is correct but the sample is rejected due to high uncertainty (or low confidence), Incorrect Accepted (IA) which is
number of samples for which the prediction is incorrect and yet the sample is accepted and Incorrect Rejected (IR) which is 
number of samples for which the prediction is incorrect and the sample is rejected often a desirable outcome in scenarios where errors should be quarantined or flagged for further review. Formally, the formulae are given below.

\begin{figure}
    \centering
    \includegraphics[width=\linewidth]{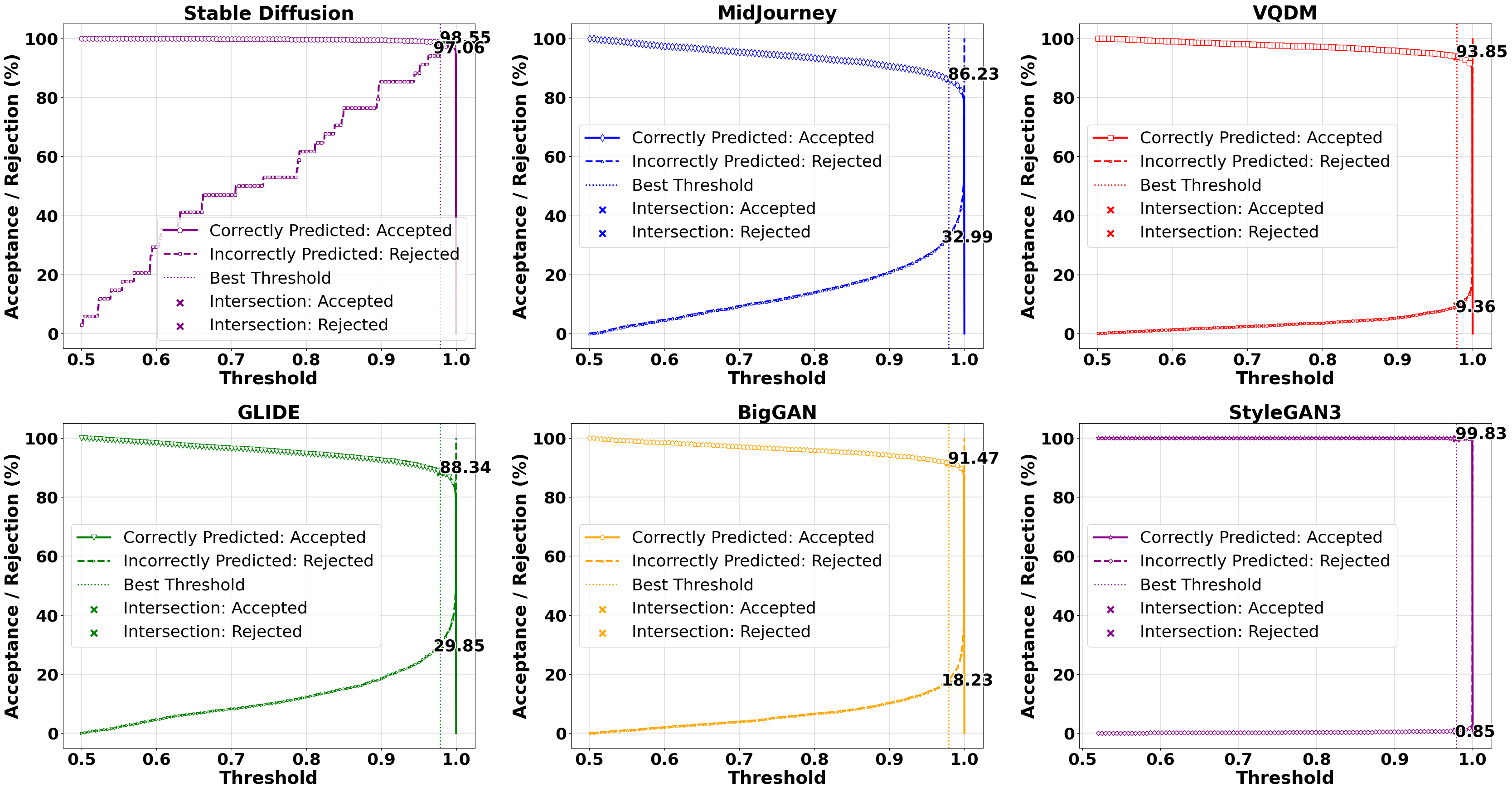}
    \caption{Correctly predicted acceptance rates and incorrectly predicted rejection rates across varying classifier probability thresholds, including the threshold \(\tau^*\) obtained using Stable Diffusion on ResNet-50.}
    \label{fig:ar_prob_resnet}
\end{figure}

\begin{figure}
    \centering
    \includegraphics[width=\linewidth]{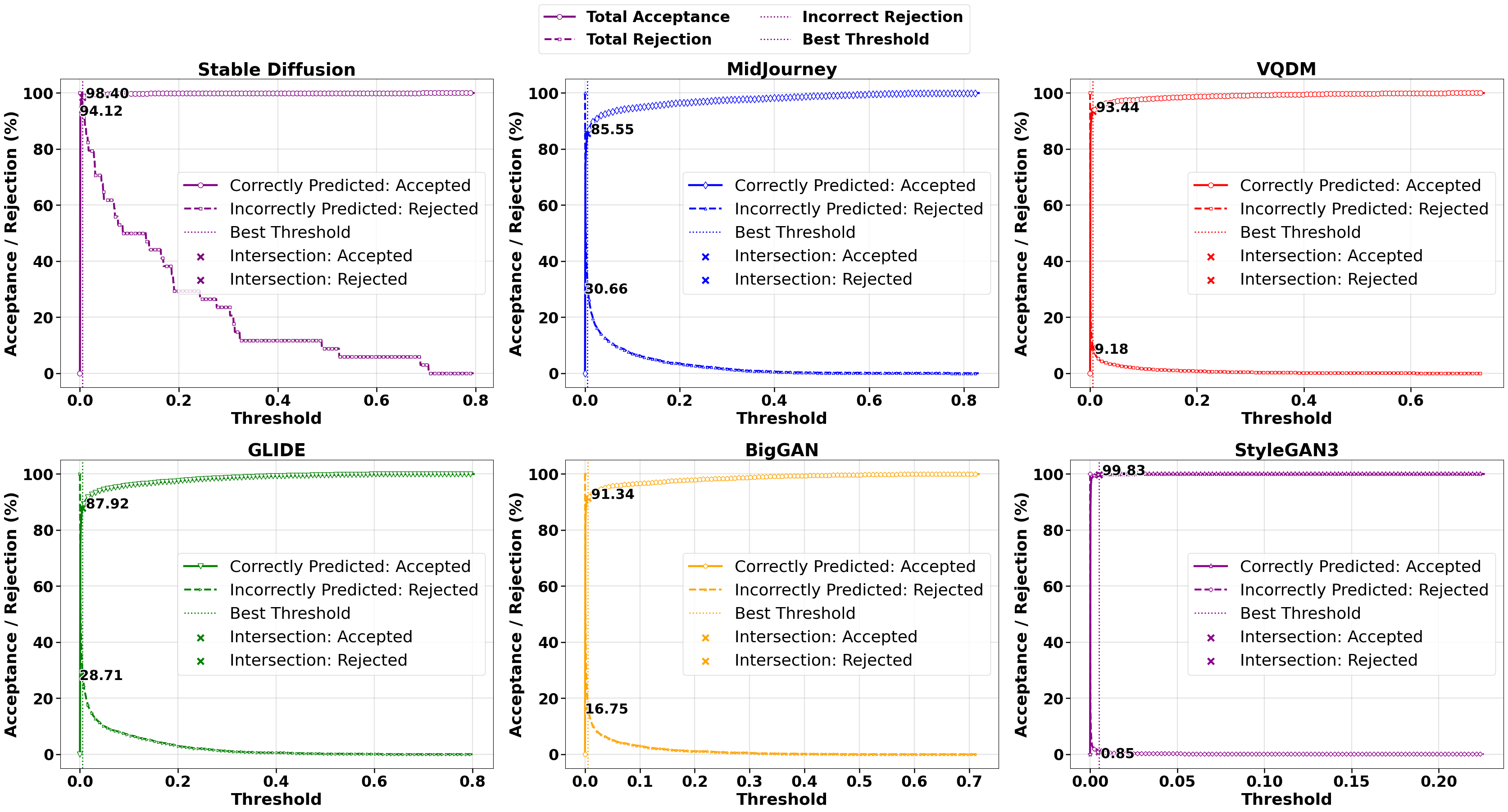}
    \caption{Correctly predicted acceptance rates and incorrectly predicted rejection rates across varying classifier total Fisher Information thresholds, including the optimal threshold \(\tau^*\) obtained using Stable Diffusion on ResNet-50.}
    \label{fig:a_r_total_fihser_resnet}
\end{figure}

{
\setlength{\abovedisplayskip}{6pt}
\setlength{\belowdisplayskip}{6pt}
\setlength{\abovedisplayshortskip}{0pt}
\setlength{\belowdisplayshortskip}{3pt}
\setlength{\parskip}{0pt} 

\begin{equation}
\text{Correct Accepted (CA)} 
= \sum_{i=1}^{N} \mathbf{1}\Bigl[\hat{y}_i = y_i \,\wedge\, (\text{acceptance criterion})\Bigr],
\end{equation}
\begin{equation}
\text{Correct Rejected (CR)} 
= \sum_{i=1}^{N} \mathbf{1}\Bigl[\hat{y}_i = y_i \,\wedge\, (\text{rejection criterion})\Bigr],
\end{equation}
\begin{equation}
\text{Incorrect Accepted (IA)} 
= \sum_{i=1}^{N} \mathbf{1}\Bigl[\hat{y}_i \neq y_i \,\wedge\, (\text{acceptance criterion})\Bigr],
\end{equation}
\begin{equation}
\text{Incorrect Rejected (IR)} 
= \sum_{i=1}^{N} \mathbf{1}\Bigl[\hat{y}_i \neq y_i \,\wedge\, (\text{rejection criterion})\Bigr],
\end{equation}

\begin{equation}
\label{eqn:max_score_general}
\tau^* = \arg\max_{\tau} \Bigl(\text{AcceptedCorrect\%}_{\text{AI}} + \text{RejectedIncorrect\%}_{\text{AI}}
+ \text{AcceptedCorrect\%}_{\text{Nature}} + \text{RejectedIncorrect\%}_{\text{Nature}}\Bigr).
\end{equation}
}
Where \(\mathbf{1}[\cdot]\) is the indicator function returning 1 if the condition is true and 0 otherwise. The acceptance criterion depends on whether \(u_i\) is treated as confidence or uncertainty, for example \((u_i \le \tau)\) vs. \((u_i \ge \tau)\).For each class of interest, such as AI (\(y_i = 0\)) or Nature (\(y_i = 1\)), these four outcomes can be computed separately by restricting the sums to samples in that class.
To determine the best threshold \(\tau\), we optimised Equation \ref{eqn:max_score_general} by performing an exhaustive search over \(\tau\) across a finely spaced range derived from the minimum and maximum Probabilities or Uncertainties in the Stable Diffusion test set. We identified the threshold \(\tau^*\) that maximises the model's ability to reject incorrect predictions while retaining as many correct predictions as possible.  This selection ensures a balance between retaining confidently correct predictions while rejecting uncertain and potentially incorrect predictions. Once the optimal threshold \(\tau^*\) was determined using Stable Diffusion, it was subsequently applied to other datasets, including MidJourney, GLIDE, VQDM and BigGAN, without further re-optimisation. This approach ensures a consistent evaluation framework across datasets and allows us to assess the generalisation of the acceptance-rejection mechanism under different data distributions. Figures \ref{fig:ar_prob_resnet} and \ref{fig:a_r_total_fihser_resnet} illustrate the correctly predicted acceptance rates and incorrectly predicted rejection rates across varying classifier probability thresholds and total Fisher Information, respectively, as an example. These plots highlight the impact of different thresholds on classification performance, with the optimal threshold \(\tau^*\) obtained using the Stable Diffusion dataset on ResNet-50. The remaining graphs for other uncertainty measures are not shown due to space constraints, but the CPA and IPR obtained on the threshold for each uncertainty measure are summarised in Table \ref{tab:resnet50_vit_cpa_ipr} due to space constraints. Figure \ref{fig:cpa_ipr} shows the scatter plot of the trade-off between CPA and IPR, where the values are computed as averages over the Midjourney, Glide, VQDM, BigGAN and SyleGan3 datasets. In this plot, each point coloured uniquely according to its corresponding uncertainty measure depicts its average CPA (i.e., the percentage of correct predictions accepted) on the x-axis and its average IPR (i.e., the percentage of incorrect predictions rejected) on the y-axis. Ideally, an uncertainty measure would achieve high CPA and high IPR (i.e., cluster near the optimal performance region of (100, 100)), indicating that it retains the correct predictions while effectively rejecting the most incorrect ones.

\begin{table}[ht!]
\centering
\small
\begin{threeparttable}
\caption{Comparison of Correctly Predicted‐Accepted (CPA) and Incorrectly Predicted‐Rejected (IPR) for ResNet50 and ViT across various datasets and uncertainty measures.}
\label{tab:resnet50_vit_cpa_ipr}
\begin{tabular}{llrrrr}
\toprule
\multirow{2}{*}{\textbf{Dataset}} & \multirow{2}{*}{\textbf{Uncertainty Measure}} 
  & \multicolumn{2}{c}{\textbf{ResNet50}} & \multicolumn{2}{c}{\textbf{ViT}} \\
\cmidrule(lr){3-4}\cmidrule(lr){5-6}
 & & \textbf{CPA} & \textbf{IPR} & \textbf{CPA} & \textbf{IPR} \\
\midrule

\multirow{8}{*}{\textbf{Stable Diffusion}} 
 & Probability               & 98.55 & 97.06 & 77.39 & 94.41 \\
 & Total Fisher Information  & 98.40 & 94.12 & 89.45 & 84.73 \\
 & Fisher Frobenius Norm     & 98.30 & 97.06 & 85.43 & 88.17 \\
 & Fisher Entropy            & 85.38 & 88.24 & 90.45 & 73.76 \\
 & Gaussian Process          & 97.04 & 97.06 & 86.91 & 92.26 \\
 & Entropy of Expected       & 20.76 & 100.00& 0.00  & 100.00 \\
 & Knowledge Uncertainty     & 22.84 & 100.00& 0.00  & 100.00 \\
 & Combined Uncertainty      & 66.93 & 100.00& 88.78 & 93.55 \\
\midrule

\multirow{8}{*}{\textbf{MidJourney}} 
 & Probability               & 86.23 & 32.99 & 68.46 & 51.06 \\
 & Total Fisher Information  & 85.55 & 30.66 & 77.76 & 32.57 \\
 & Fisher Frobenius Norm     & 85.09 & 31.74 & 75.04 & 36.12 \\
 & Fisher Entropy            & 76.08 & 29.61 & 80.33 & 37.51 \\
 & Gaussian Process          & 80.54 & 40.50 & 72.93 & 58.70 \\
 & Entropy of Expected       & 20.65 & 99.36 & 0.00  & 100.00 \\
 & Knowledge Uncertainty     & 18.73 & 99.51 & 0.00  & 100.00 \\
 & Combined Uncertainty      & 51.89 & 87.42 & 74.75 & 52.54 \\
\midrule

\multirow{8}{*}{\textbf{VQDM}} 
 & Probability               & 93.85 & 9.36  & 77.71 & 28.63 \\
 & Total Fisher Information  & 93.44 & 9.18  & 85.21 & 16.70 \\
 & Fisher Frobenius Norm     & 93.59 & 9.12  & 83.72 & 18.41 \\
 & Fisher Entropy            & 88.89 & 10.16 & 85.21 & 19.78 \\
 & Gaussian Process          & 89.91 & 18.79 & 79.03 & 30.52 \\
 & Entropy of Expected       & 23.75 & 79.88 & 0.06  & 99.98 \\
 & Knowledge Uncertainty     & 26.85 & 78.19 & 0.05  & 99.98 \\
 & Combined Uncertainty      & 61.62 & 54.51 & 81.52 & 26.27 \\
\midrule

\multirow{8}{*}{\textbf{GLIDE}} 
 & Probability               & 88.34 & 29.85 & 69.98 & 48.71 \\
 & Total Fisher Information  & 87.92 & 28.71 & 78.16 & 32.65 \\
 & Fisher Frobenius Norm     & 87.81 & 29.28 & 76.18 & 34.92 \\
 & Fisher Entropy            & 74.82 & 30.29 & 79.80 & 37.45 \\
 & Gaussian Process          & 85.52 & 42.01 & 72.43 & 56.66 \\
 & Entropy of Expected       & 17.98 & 99.59 & 0.00  & 100.00 \\
 & Knowledge Uncertainty     & 16.81 & 99.59 & 0.00  & 100.00 \\
 & Combined Uncertainty      & 49.41 & 87.90 & 74.66 & 50.40 \\
\midrule

\multirow{8}{*}{\textbf{BigGAN}} 
 & Probability               & 91.47 & 18.23 & 82.46 & 24.40 \\
 & Total Fisher Information  & 91.34 & 16.75 & 88.97 & 15.26 \\
 & Fisher Frobenius Norm     & 91.28 & 17.72 & 88.09 & 16.32 \\
 & Fisher Entropy            & 88.72 & 18.19 & 87.07 & 17.15 \\
 & Gaussian Process          & 88.50 & 25.59 & 82.99 & 25.51 \\
 & Entropy of Expected       & 23.63 & 99.87 & 0.08  & 99.91 \\
 & Knowledge Uncertainty     & 23.42 & 99.85 & 0.06  & 99.93 \\
 & Combined Uncertainty      & 61.40 & 92.67 & 84.76 & 21.61 \\
 \midrule
 \multirow{8}{*}{\textbf{StyleGAN3}}
 & Probability               & 99.83 & 0.85  & 51.63 & 55.40 \\
 & Total Fisher Information  & 99.83 & 0.85  & 65.98 & 33.36 \\
 & Fisher Frobenius Norm     & 99.83 & 0.85  & 63.67 & 37.97 \\
 & Fisher Entropy            & 99.83 & 2.32  & 63.03 & 40.13 \\
 & Gaussian Process          & 99.32 & 12.61 & 44.09 & 64.04 \\
 & Entropy of Expected       & 17.66 & 58.38 & 0.00  & 100.00 \\
 & Knowledge Uncertainty     & 12.73 & 62.25 & 0.00  & 100.00 \\
 & Combined Uncertainty      & 85.48 & 29.08 & 51.39 & 54.95 \\

\bottomrule
\end{tabular}
\end{threeparttable}
\end{table}

\begin{table}[H]
\centering
\small      
\begin{threeparttable}
\caption{Rejection rates (Total, Correct Prediction, and Incorrect Prediction) for various confidence and uncertainty measures when ResNet50, trained on Stable Diffusion, is tested on Stable Diffusion (SD), Midjourney (MJ), GLIDE (GD), VQDM (VQ),BigGAN (BG) and StyleGAN3 (SG)}
\label{tab:ai_nature_rejection}
\begin{tabular}{ll rrr rrr}
\toprule
\multirow{2}{*}{\textbf{Measure}} & \multirow{2}{*}{\textbf{Dataset}} & \multicolumn{3}{c}{\textbf{Nature}} & \multicolumn{3}{c}{\textbf{AI}} \\
\cmidrule(lr){3-5} \cmidrule(lr){6-8}
 & & \textbf{Total} & \textbf{Correct} & \textbf{Incorrect} & \textbf{Total} & \textbf{Correct} & \textbf{Incorrect} \\
\midrule

\multirow{5}{*}{\textbf{Probability}} 
 & SD  & 1.45 & 1.28 & 93.33 & 1.85 & 1.62 & 100.00 \\
 & MJ  & 1.50 & 1.25 & 88.24 & 39.10 & 52.37 & 32.76 \\
 & GD  & 1.70 & 1.49 & 92.86 & 31.22 & 33.04 & 29.57 \\
 & VQ  & 1.50 & 1.22 & 80.95 & 13.55 & 39.61 & 9.06 \\
 & BG  & 1.47 & 1.29 & 73.33 & 24.13 & 67.76 & 18.07 \\
  & SG  & 0.17 & 0.17 & 0.00  & 0.85 & 0.00 & 0.85 \\
\midrule

\multirow{5}{*}{\textbf{Total Fisher Information}} 
 & SD  & 1.40 & 1.23 & 93.33 & 2.20 & 1.98 & 94.74 \\
 & MJ  & 1.43 & 1.19 & 88.24 & 38.48 & 55.36 & 30.42 \\
 & GD  & 1.60 & 1.39 & 92.86 & 31.33 & 34.55 & 28.43 \\
 & VQ  & 1.48 & 1.19 & 85.71 & 13.88 & 43.02 & 8.87 \\
 & BG  & 1.42 & 1.20 & 86.67 & 23.03 & 69.67 & 16.55 \\
& SG  & 0.17 & 0.17 & 0.00  & 0.85 & 0.00 & 0.85 \\
\midrule

\multirow{5}{*}{\textbf{Fisher Frobenius Norm}} 
 & SD  & 1.43 & 1.24 & 100.00 & 2.39 & 2.17 & 94.74 \\
 & MJ  & 1.53 & 1.27 & 94.12 & 39.72 & 56.96 & 31.48 \\
 & GD  & 1.65 & 1.44 & 92.86 & 31.75 & 34.80 & 29.00 \\
 & VQ  & 1.48 & 1.19 & 85.71 & 13.67 & 41.88 & 8.81 \\
 & BG  & 1.48 & 1.27 & 86.67 & 23.88 & 69.67 & 17.52 \\
  & SG  & 0.17 & 0.17 & 0.00  & 0.85 & 0.00 & 0.85 \\
\midrule

\multirow{5}{*}{\textbf{Fisher Entropy}} 
 & SD  & 1.19 & 1.00 & 100.00 & 28.36 & 28.24 & 78.95 \\
 & MJ  & 1.45 & 1.17 & 100.00 & 50.25 & 94.07 & 29.31 \\
 & GD  & 1.30 & 1.07 & 100.00 & 51.75 & 75.84 & 29.98 \\
 & VQ  & 1.57 & 1.22 & 100.00 & 19.83 & 78.21 & 9.79 \\
 & BG  & 1.30 & 1.05 & 100.00 & 27.35 & 94.95 & 17.96 \\
 & SG  & 0.17 & 0.17 & 0.00  & 2.32 & 0.00 & 2.32 \\
\midrule

\multirow{5}{*}{\textbf{Gaussian Process}} 
 & SD  & 4.25 & 4.08 & 93.33 & 2.08 & 1.84 & 100.00 \\
 & MJ  & 5.45 & 5.18 & 100.00 & 47.77 & 63.51 & 40.25 \\
 & GD  & 4.80 & 4.63 & 78.57 & 38.68 & 35.18 & 41.85 \\
 & VQ  & 4.60 & 4.35 & 76.19 & 23.03 & 49.04 & 18.56 \\
 & BG  & 4.80 & 4.59 & 86.67 & 30.82 & 78.14 & 24.24 \\
  & SG  & 0.68 & 0.68 & 0.00  & 12.61 & 0.00 & 12.61 \\
\midrule

\multirow{5}{*}{\textbf{Entropy of Expected}} 
 & SD  & 74.06 & 74.01 & 100.00 & 84.50 & 84.46 & 100.00 \\
 & MJ  & 72.75 & 72.67 & 100.00 & 99.55 & 99.95 & 99.36 \\
 & GD  & 73.57 & 73.50 & 100.00 & 99.75 & 99.93 & 99.59 \\
 & VQ  & 72.92 & 72.82 & 100.00 & 82.70 & 99.55 & 79.80 \\
 & BG  & 73.55 & 73.48 & 100.00 & 99.88 & 100.00 & 99.87 \\
  & SG  & 82.34 & 82.34 & 0.00  & 58.38 & 0.00  & 58.38 \\
\midrule

\multirow{5}{*}{\textbf{Knowledge Uncertainty}} 
 & SD  & 72.47 & 72.42 & 100.00 & 81.94 & 81.89 & 100.00 \\
 & MJ  & 75.28 & 75.21 & 100.00 & 99.65 & 99.95 & 99.51 \\
 & GD  & 75.28 & 75.23 & 100.00 & 99.75 & 99.93 & 99.59 \\
 & VQ  & 69.47 & 69.38 & 95.24 & 81.15 & 98.75 & 78.12 \\

 & BG  & 73.78 & 73.72 & 100.00 & 99.87 & 100.00 & 99.85 \\
   & SG  & 87.27 & 87.27 & 0.00  & 62.25 & 0.00 & 62.25 \\
\midrule

\multirow{5}{*}{\textbf{Combined Uncertainty}} 
 & SD  & 32.51 & 32.39 & 100.00 & 33.91 & 33.76 & 100.00 \\
 & MJ  & 32.35 & 32.16 & 100.00 & 90.58 & 97.32 & 87.36 \\
 & GD  & 32.12 & 31.96 & 100.00 & 88.75 & 89.75 & 87.85 \\
 & VQ  & 31.43 & 31.19 & 100.00 & 59.15 & 87.17 & 54.33 \\
 & BG  & 31.38 & 31.21 & 100.00 & 93.43 & 99.04 & 92.65 \\
  & SG  & 14.52 & 14.52 & 0.00  & 29.08 & 0.00 & 29.08 \\
 
\bottomrule
\end{tabular}
\end{threeparttable}
\end{table}

\begin{table}[H]
\centering
\begin{threeparttable}
\caption{Rejection rates (Total, Correct Prediction, and Incorrect Prediction) for various confidence and uncertainty measures when ViT, trained on Stable Diffusion, is tested on Stable Diffusion (SD), MidJourney (MJ), GLIDE (GD), VQDM (VQ), and BigGAN (BG).}
\label{tab:vit_reject}
\begin{tabular}{ll rrr rrr}
\toprule
\multirow{2}{*}{\textbf{Measure}} & \multirow{2}{*}{\textbf{Dataset}} & \multicolumn{3}{c}{\textbf{Nature}} & \multicolumn{3}{c}{\textbf{AI}} \\
\cmidrule(lr){3-5} \cmidrule(lr){6-8}
 & & \textbf{Total} & \textbf{Correct} & \textbf{Incorrect} & \textbf{Total} & \textbf{Correct} & \textbf{Incorrect} \\
\midrule

\multirow{5}{*}{\textbf{Probability}} 
 & SD  & 15.40 & 12.62 & 95.52 & 34.00 & 32.51 & 92.89 \\
 & MJ  & 16.05 & 13.28 & 97.96 & 60.87 & 86.06 & 48.79 \\
 & GD  & 15.75 & 12.89 & 96.59 & 58.68 & 92.62 & 46.49 \\
 & VQ  & 15.47 & 13.12 & 95.88 & 34.85 & 93.60 & 26.46 \\
 & BG  & 15.47 & 12.64 & 97.50 & 26.30 & 97.47 & 21.81 \\
  & SG  & 45.08 & 38.75 & 98.40 & 57.75 & 98.51 & 50.99 \\
\midrule

\multirow{5}{*}{\textbf{Total Fisher Information}} 
 & SD  & 8.43 & 5.69 & 87.31 & 16.99 & 15.37 & 81.22 \\
 & MJ  & 9.37 & 6.69 & 88.78 & 42.43 & 68.67 & 29.86 \\
 & GD  & 9.05 & 6.18 & 90.24 & 42.95 & 79.07 & 29.97 \\
 & VQ  & 8.45 & 6.23 & 84.71 & 22.85 & 81.33 & 14.50 \\
 & BG  & 8.95 & 6.16 & 90.00 & 17.23 & 90.45 & 12.62 \\
  & SG  & 30.56 & 23.55 & 89.60 & 36.27 & 88.61 & 27.59 \\

\midrule

\multirow{5}{*}{\textbf{Fisher Frobenius Norm}} 
 & SD  & 9.56 & 6.71 & 91.79 & 23.85 & 22.35 & 83.25 \\
 & MJ  & 10.47 & 7.68 & 92.86 & 47.37 & 76.54 & 33.38 \\
 & GD  & 10.10 & 7.16 & 93.17 & 46.08 & 84.68 & 32.22 \\
 & VQ  & 9.47 & 7.12 & 90.00 & 25.02 & 87.47 & 16.10 \\
 & BG & 9.80 & 6.98 & 91.50 & 18.32 & 92.13 & 13.66 \\
  & SG  & 32.43 & 25.26 & 92.80 & 41.13 & 94.06 \\
\midrule

\multirow{5}{*}{\textbf{Fisher Entropy}} 
 & SD  & 10.44 & 8.43 & 68.28 & 12.39 & 10.65 & 81.22 \\
 & MJ  & 10.92 & 8.92 & 69.90 & 41.07 & 51.75 & 35.95 \\
 & GD  & 10.23 & 8.28 & 65.37 & 43.45 & 63.75 & 36.16 \\
 & VQ  & 10.65 & 9.02 & 66.47 & 23.43 & 59.60 & 18.27 \\
 & BG  & 10.72 & 8.71 & 69.00 & 19.25 & 81.74 & 15.31 \\
  & SG  & 37.18 & 31.62 & 84.00 & 39.79 & 64.85 & 35.63 \\
\midrule

\multirow{5}{*}{\textbf{Gaussian Process}} 
 & SD  & 15.25 & 12.58 & 92.16 & 15.53 & 13.58 & 92.39 \\
 & MJ  & 15.75 & 13.15 & 92.86 & 60.80 & 68.62 & 57.05 \\
 & GD  & 15.27 & 12.58 & 91.22 & 62.27 & 82.35 & 55.05 \\
 & VQ  & 14.98 & 12.81 & 89.41 & 35.58 & 84.40 & 28.61 \\
 & BG  & 15.18 & 12.48 & 93.50 & 27.15 & 95.22 & 22.86 \\
  & SG  & 54.67 & 50.65 & 98.00 & 64.79 & 93.96 & 61.37 \\
\midrule

\multirow{5}{*}{\textbf{Entropy of Expected}} 
 & SD  & 100.00 & 100.00 & 100.00 & 100.00 & 100.00 & 100.00 \\
 & MJ  & 100.00 & 100.00 & 100.00 & 100.00 & 100.00 & 100.00 \\
 & GD  & 100.00 & 100.00 & 100.00 & 100.00 & 100.00 & 100.00 \\
 & VQ  & 99.93 & 99.93 & 100.00 & 99.98 & 100.00 & 99.98 \\
 & BG  & 99.92 & 99.91 & 100.00 & 99.92 & 100.00 & 99.91 \\
& SG  & 100.00 & 100.00 & 100.00 & 100.00 & 100.00 & 100.00 \\
\midrule

\multirow{5}{*}{\textbf{Knowledge Uncertainty}} 
 & SD  & 100.00 & 100.00 & 100.00 & 100.00 & 100.00 & 100.00 \\
 & MJ  & 100.00 & 100.00 & 100.00 & 100.00 & 100.00 & 100.00 \\
 & GD  & 100.00 & 100.00 & 100.00 & 100.00 & 100.00 & 100.00 \\
 & VQ  & 99.95 & 99.97 & 99.41 & 99.98 & 99.87 & 100.00 \\
 & BG  & 99.93 & 99.95 & 99.50 & 99.93 & 99.72 & 99.95 \\
  & SG  & 100.00 & 100.00 & 100.00 & 100.00 & 100.00 & 100.00\\
\midrule

\multirow{5}{*}{\textbf{Combined Uncertainty}} 
 & SD  & 13.00 & 10.24 & 92.54 & 14.22 & 12.19 & 94.92 \\
 & MJ  & 13.45 & 10.84 & 90.82 & 56.38 & 68.26 & 50.69 \\
 & GD  & 12.92 & 10.15 & 91.22 & 57.05 & 80.83 & 48.50 \\
 & VQ  & 12.43 & 10.14 & 91.18 & 31.57 & 83.33 & 24.17 \\
 & BG  & 13.13 & 10.34 & 94.00 & 23.55 & 94.94 & 19.05 \\
  & SG  & 45.93 & 40.08 & 95.20 & 56.83 & 93.07 & 50.82 \\
\bottomrule
\end{tabular}
\end{threeparttable}
\end{table}

One of the main aims of our analysis is to examine the total rejection rate, correct rejection rate, and incorrect rejection rate separately for the AI and Nature classes. Under conditions of dataset shift, misclassifications are concentrated in the AI class, making it particularly important to detect and reject AI‐generated errors before they can be mistakenly accepted as Natural images. By breaking down these rates for AI versus Nature, we can pinpoint where errors occur, how many are correctly intercepted, and whether the system is discarding valid samples at an acceptable level. We define these rates as follows. The total rejection rate measures the fraction of all samples, whether correct or incorrect, that end up being rejected:

\begin{equation}
 \text{Total Rejection Rate} = 100 \times \frac{CR + IR}{CA + CR + IA + IR},   
\end{equation}

where \(CA\) denotes Correct Accepted, \(CR\) Correct Rejected, \(IA\) Incorrect Accepted, and \(IR\) Incorrect Rejected. For the correctly classified samples, the correct rejection rate specifies the percentage of those that still undergo rejection.

\begin{equation}
  \text{Correct Rejection Rate} = 100 \times \frac{CR}{CA + CR},  
\end{equation}

Among the misclassified samples, the incorrect rejection rate highlights how many of those errors are caught and flagged by the model:

\begin{equation}
 \text{Incorrect Rejection Rate} = 100 \times \frac{IR}{IA + IR}.   
\end{equation}

Each of these rates can be computed separately for AI and Nature by restricting \((CA, CR, IA, IR)\) to instances belonging to the respective label. To capture an overall sense of how strict the mechanism is, we look at the total rejection rate, which indicates what fraction of all samples, correct or not, are flagged. The correct rejection rate then shows how many legitimate predictions are needlessly filtered out, shedding light on whether the threshold might be too conservative. Finally, the incorrect rejection rate reveals how effectively actual misclassifications are quarantined.  

\begin{figure}[h] 
    \centering
    \includegraphics[width=\linewidth]{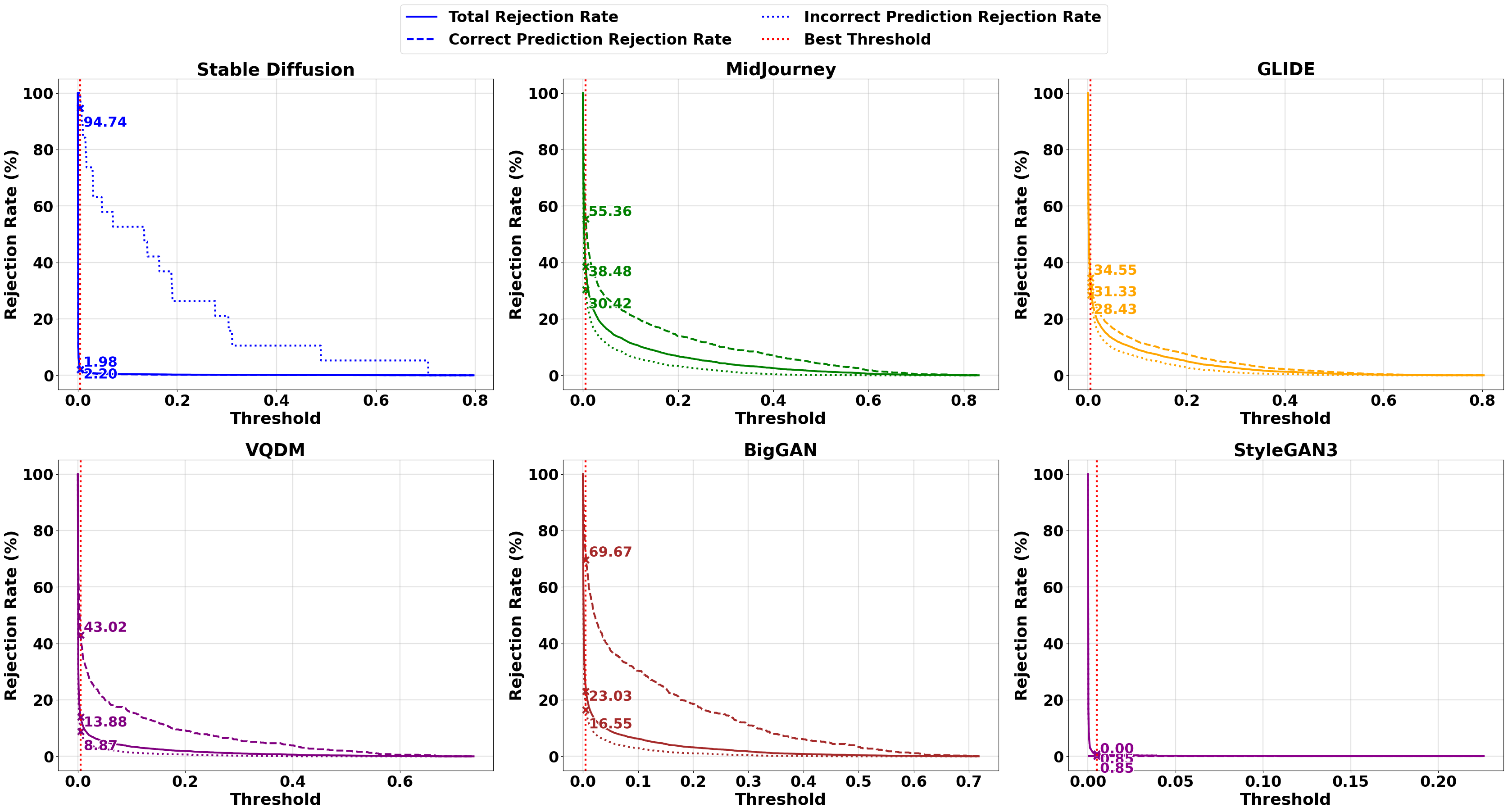}
    \caption{Rejection rates for the AI class (Total, Correct Prediction, and Incorrect Prediction) using the optimal threshold \(\tau^*\) based on Total Fisher Information, derived from ResNet50 trained on Stable Diffusion.}
    \label{fig:ai_total_fisher}
\end{figure}

\begin{figure}
    \centering
    \includegraphics[width=\linewidth]{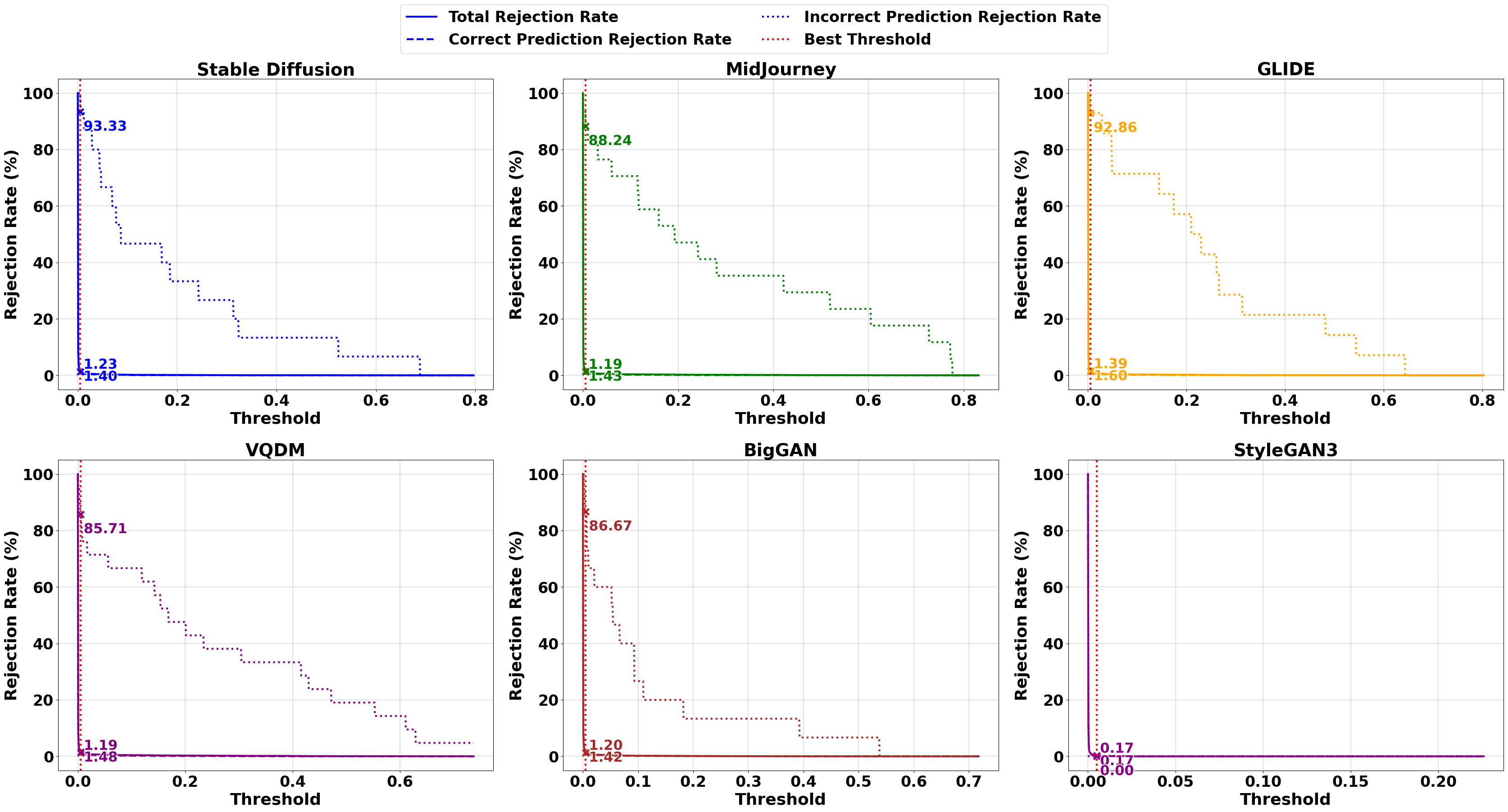}
    \caption{Rejection rates for the Nature class (Total, Correct Prediction, and Incorrect Prediction) using the optimal threshold \(\tau^*\) based on Total Fisher Information, derived from ResNet50 trained on Stable Diffusion.}
    \label{fig:nature_total_fisher}
\end{figure}

In scenarios where dataset shifts magnify AI errors, it may be desirable to tolerate a relatively high total rejection rate for AI, even if it includes some correctly predicted AI samples, if doing so substantially reduces the risk of synthetic content slipping through. This trade‐off can be acceptable as long as the acceptance of correctly predicted Nature samples and In-domain AI images is not unduly compromised. As an example Figures \ref{fig:ai_total_fisher} and \ref{fig:nature_total_fisher}
Illustrate the AI and Nature rejection rates, respectively, under Total Fisher Information for ResNet‐50 using the best threshold obtained from Stable Diffusion. We focus on ResNet‐50 due to its superior performance (in terms of accuracy, precision, recall, and F1 score) compared to ViT, providing a more reliable assessment of how the acceptance-rejection mechanism handles data distribution shifts. Due to space limitations, the remaining rejection rates computed at this threshold are presented in Table \ref{tab:ai_nature_rejection}. While the Vision Transformer (ViT) was also evaluated as part of our broader study, its slightly lower baseline performance makes ResNet‐50 a more illustrative example. As a result, our discussion primarily focuses on ResNet‐50, with ViT results included for completeness. Table \ref{tab:vit_reject} presents the rejection rates obtained using the ViT model.

To further illustrate this trade-off, Figures \ref{fig:ai_scatter_plot} and \ref{fig:nature_scatter_plot} present scatter plots that compare the average Correct Prediction Rejection Rate (CPRR) and Incorrect Prediction Rejection Rate (IPRR) computed over the five non–Stable Diffusion datasets (Midjourney, Glide, VQDM, BigGAN and StyleGAN3). In these plots, each point is coloured uniquely according to its corresponding uncertainty measure, with the x-axis indicating the average CPRR (i.e., the percentage of correctly predicted samples that are mistakenly rejected) and the y-axis representing the average IPRR (i.e., the percentage of incorrect predictions that are successfully rejected). The variations in generation methods can lead to shifts in the data distribution, making AI classification more challenging as shown in Table \ref{tab:conf-matrix-data-shift-res} and \ref{tab:conf-matrix-data-shift-vit}; hence, a more conservative rejection mechanism (i.e., a higher CPRR) is sometimes justified to ensure that incorrect predictions are effectively filtered out. 

\subsubsection*{Detailed Analysis and Interpretation of Results}

The results in Table \ref{tab:resnet50_vit_cpa_ipr} present a detailed comparison of Correctly Predicted Accepted (CPA) and Incorrectly Predicted Rejected (IPR) rates for ResNet50 across different uncertainty measures and datasets. On the in-distribution dataset, Stable Diffusion, the model performs exceptionally well across most uncertainty metrics. Probability, Total Fisher Information, and Fisher Frobenius Norm each achieve CPA rates above 98\% and IPR rates exceeding 94\%, indicating strong reliability in both accepting correct and rejecting incorrect predictions. Fisher Entropy, while slightly less conservative, still maintains a good balance. Gaussian Process uncertainty also performs strongly here. However, Entropy of Expected and Knowledge Uncertainty, though achieving perfect IPR (100\%), do so at the cost of drastically reduced CPA, rejecting most correct predictions. The Combined Uncertainty measure offers a more balanced alternative, accepting around 67\% of correct predictions while still rejecting all incorrect ones.

\begin{figure}[H]
    \centering
    \includegraphics[width=\linewidth]{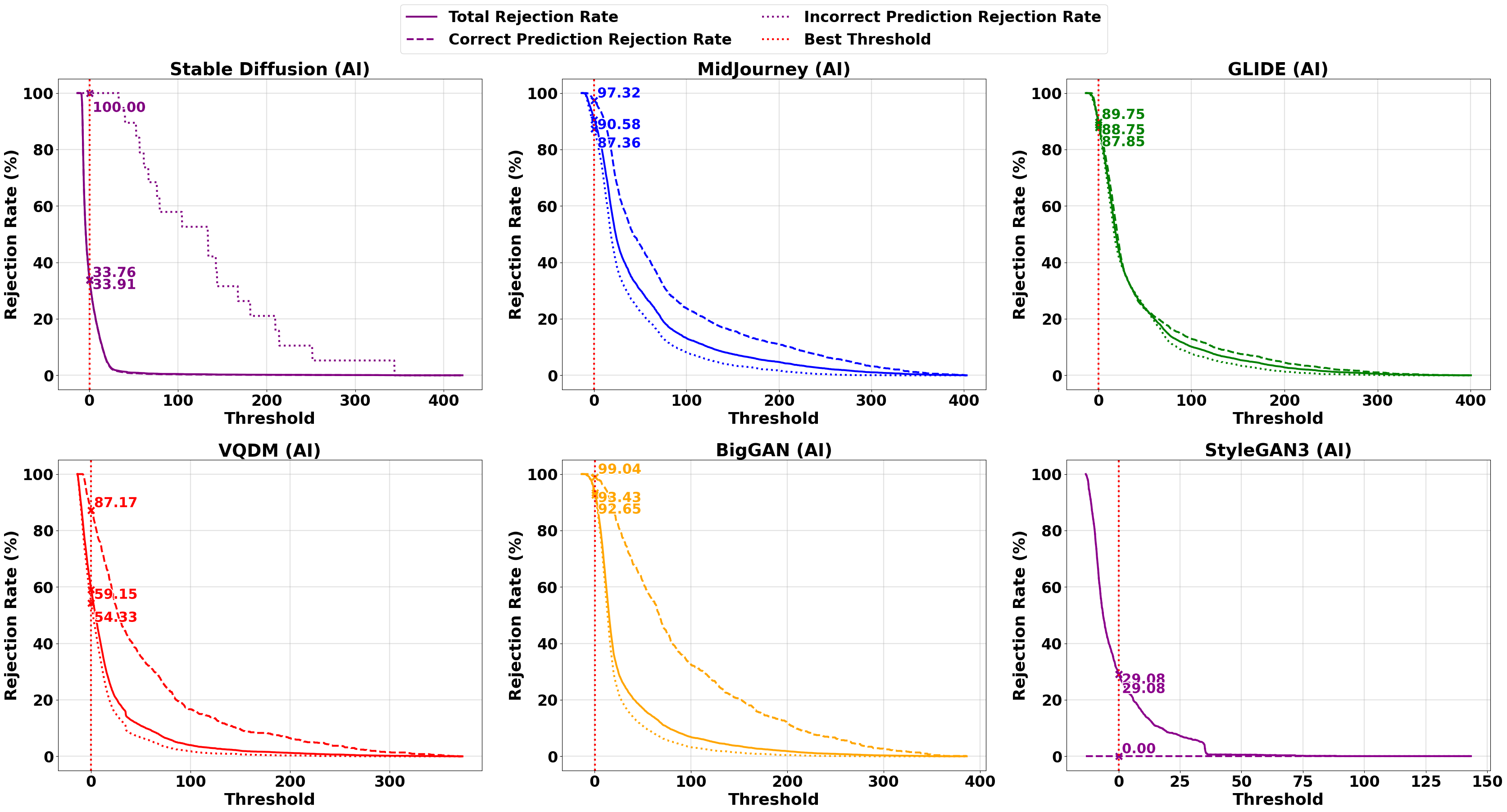}
    \caption{Rejection rates for the AI class (Total, Correct Prediction, and Incorrect Prediction) using the optimal threshold \(\tau^*\) based on Combine Uncertainty, derived from ResNet50 trained on Stable Diffusion.}
    \label{fig:ai_combined}
\end{figure}

\begin{figure}[H]
    \centering
    \includegraphics[width=\linewidth]{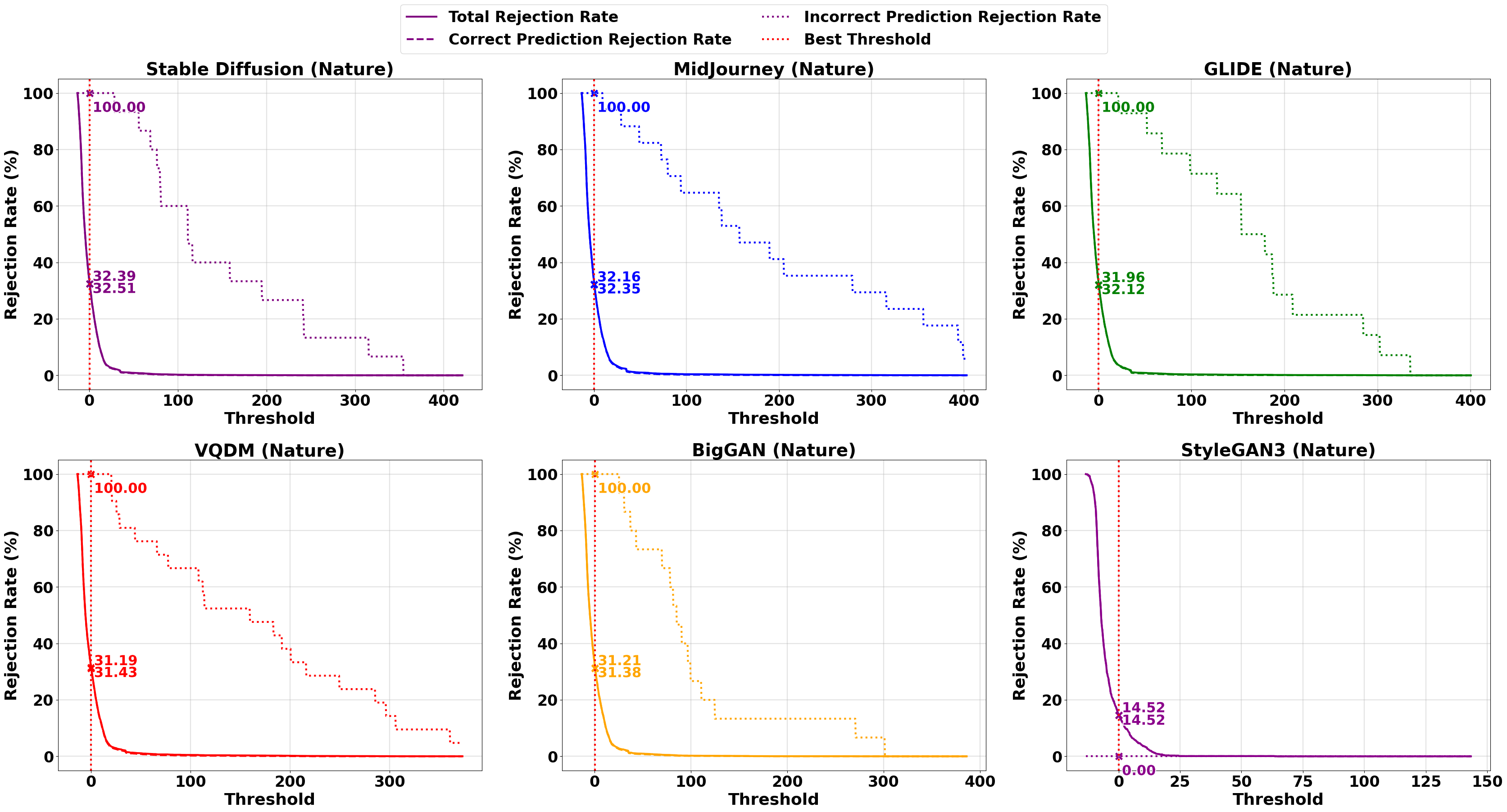}
    \caption{Rejection rates for the Nature class (Total, Correct Prediction, and Incorrect Prediction) using the optimal threshold \(\tau^*\) based on  Combine Uncertainty, derived from ResNet50 trained on Stable Diffusion.}
    \label{fig:nature_combined}
\end{figure}

\begin{figure}[H]
    \centering
        \begin{minipage}{0.48\textwidth}
        \centering
        \includegraphics[width=\linewidth]{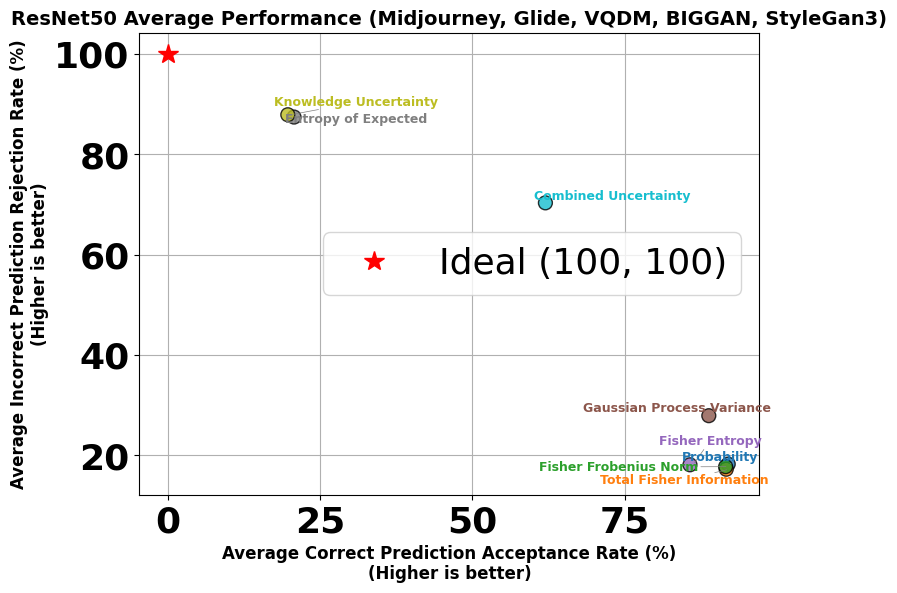}
        \caption{Trade-off between average Correct Prediction Acceptance (CPA) versus Incorrect Prediction Rejection (IPR).}
        \label{fig:cpa_ipr}
    \end{minipage}
    \vspace{0.5cm}
    \vspace{0.5cm}
    \begin{minipage}{0.48\textwidth}
        \centering
        \includegraphics[width=\linewidth]{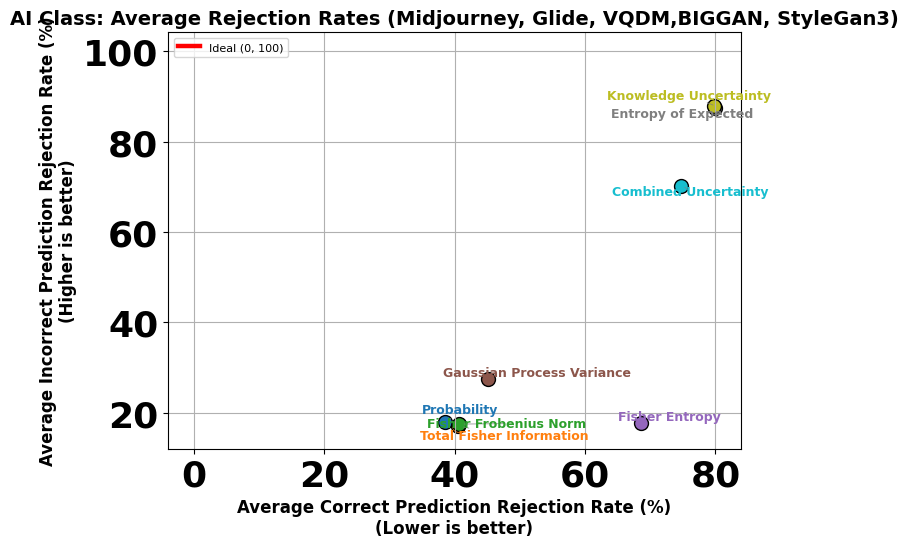}
        \caption{Trade-off between average Correct Prediction Rejection Rate (CPRR) versus Incorrect Prediction Rejection Rate (IPRR) for the \textbf{AI} class.}
        \label{fig:ai_scatter_plot}
    \end{minipage}%
    \hspace{0.1cm}
    \begin{minipage}{0.48\textwidth}
        \centering
        \includegraphics[width=\linewidth]{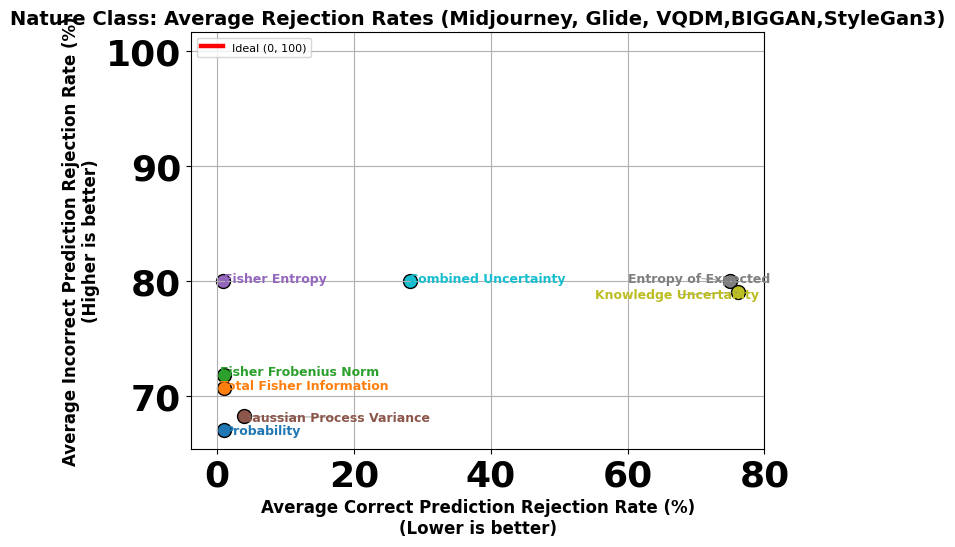}
        \caption{Trade-off between average Correct Prediction Rejection Rate (CPRR) versus Incorrect Prediction Rejection Rate (IPRR) for the \textbf{Nature} class.}
        \label{fig:nature_scatter_plot}
    \end{minipage}
    

\end{figure}

Under distributional shift evaluated on Midjourney, GLIDE, VQDM, BigGAN and StyleGAN3, the performance of individual Uncertainty measures declines, particularly in rejecting incorrect predictions. For instance, in Midjourney, while CPA remains above 85\% for probability and Fisher-based methods, their IPR drops below 33\%. In contrast, the Gaussian Process measure demonstrates slightly better robustness under shift, improving IPR by 5–10 percentage points compared to probability or Fisher metrics, though still far from ideal. Entropy of Expected and Knowledge Uncertainty behave differently. Across all out-of-distribution datasets, they consistently achieve near-perfect IPRs, successfully rejecting almost all incorrect predictions. However, they also reject a significant fraction of correct ones—CPA falls below 25\% in most cases. This overly conservative behaviour limits their practicality in settings where preserving accurate predictions is critical.
Combined Uncertainty emerges as the most robust and balanced approach across datasets. While its CPA under shift is lower than in-distribution (e.g., around 52\% on Midjourney, 49\% on GLIDE, 62\% on VQDM, 61\% on BigGAN and 85\% on StyleGAN3 ), it compensates with significantly higher IPR, ranging from 29\% to 93\%, outperforming all other individual metrics in rejecting incorrect predictions. This trade-off of rejecting more correct predictions to prevent misclassification may be acceptable in high-risk applications, mainly since rejected synthetic samples can be used for further retraining.

The trade-offs become even clearer when these metrics are visualised as a scatter plot in Figure \ref{fig:cpa_ipr}, which is generated from the averaged data over Midjourney, Glide, VQDM and BigGAN.  The red star at (100, 100) denotes the ideal performance, perfectly accepting all correct predictions while rejecting all incorrect ones. Under dataset shift, however, most measures fail to reach this ideal. For instance, measures such as probability, Total Fisher Information, and Fisher Frobenius Norm cluster at relatively high CPA values (in the mid-80s to low-90s) but exhibit low IPR values (around 29–33\%), indicating that although they retain most correct predictions, they do not effectively reject incorrect ones. Conversely, measures based on Entropy of Expected and Knowledge Uncertainty achieve near-perfect IPR values (close to 100\%), but their CPA values are significantly lower, reflecting an overly aggressive rejection that may discard many correct predictions. As mentioned before, the Combined Uncertainty measure demonstrates a more balanced performance. It achieves a moderate CPA of approximately 60\% while delivering a high IPR of about 70\%. 

Given that most misclassifications under dataset shift occur within the AI class, adopting a more conservative rejection threshold can be justified, particularly if it helps block misclassified synthetic content while keeping performance stable for the Nature class and in-distribution AI samples. Table \ref{tab:ai_nature_rejection} illustrates this dynamic in detail.
On the in-distribution Stable Diffusion data, Uncertainty measures like probability, Total Fisher Information and Fisher Frobenius Norm perform nearly perfectly, with high Correct Prediction Acceptance (CPA) and strong Incorrect Prediction Rejection (IPR) rates across both AI and Nature classes. However, this reliable behaviour deteriorates once the model is exposed to distributional shift—as seen on Midjourney (MJ), GLIDE (GD), VQDM (VQ) BigGAN(BG) and StyleGAN3(SG). For the AI class in particular, these same measures now exhibit Correct Prediction Rejection Rates (CPRRs) ranging from 40\% to 75\% and low IPRRs between 20\% and 30\%, as visualised in Figure \ref{fig:ai_scatter_plot}. In other words, they begin rejecting many correct AI predictions while failing to filter out a significant number of incorrect ones.
On the other hand, the entropy of expectation and knowledge uncertainty behaves in a highly conservative manner. In the AI class, both measures yield CPRRs and IPRRs close to 100\%, effectively rejecting nearly every prediction, correct or incorrect. While this strategy is successful at blocking synthetic errors, it does so by indiscriminately discarding a large portion of correct predictions, particularly from the Nature class, as shown in Figure \ref{fig:nature_scatter_plot}.
The Combined Uncertainty measure, however, achieves a more practical balance. While it does have a higher CPRR in the AI class, reflecting that some correct predictions are sacrificed, it compensates with an IPRR of approximately 70\%, meaning it filters out most misclassified synthetic inputs (Figure \ref{fig:ai_scatter_plot}). In the Nature class (Figure \ref{fig:nature_scatter_plot}), Combined Uncertainty manages to retain nearly all correct predictions, showing a low CPRR while still maintaining a high rejection of incorrect ones at 80\%. This class-specific stability is crucial in scenarios where natural content must be preserved.
Figures \ref{fig:ai_combined} and \ref{fig:nature_combined} further support this strategy. When applying the optimal rejection threshold $\tau$ derived from the Combined Uncertainty signal, the AI class benefits from strong rejection of incorrect predictions, while the Nature class continues to preserve accurate predictions and reject errors effectively.

In summary, the results from Tables \ref{tab:resnet50_vit_cpa_ipr} and \ref{tab:ai_nature_rejection}, along with Figures \ref{fig:ai_scatter_plot} and \ref{fig:nature_scatter_plot}, highlight an important trade-off. While standard uncertainty measures perform well on in-distribution data, their effectiveness drops significantly under distributional shifts, particularly for the AI class. Overly conservative approaches like Entropy of Expected and Knowledge Uncertainty reject nearly all predictions, limiting their utility. In contrast, the Combined Uncertainty measure offers a more desirable middle ground. This balance is especially valuable given that most errors under shift occur in the AI class, where strict filtering of synthetic content is essential. Meanwhile, performance in the Nature class remains stable. Still, as dataset shift continues to increase and overall rejection rates rise, it may become necessary to periodically retrain the model to maintain robustness and adapt to evolving distributions.

\subsubsection*{ Adversarial Robustness Evaluation}

To evaluate the reliability and security of the proposed AI-generated image detection framework, we conducted a series of adversarial robustness experiments. We employed two widely used white-box attack methods: the Fast Gradient Sign Method (FGSM) and the Projected Gradient Descent (PGD). FGSM generates adversarial examples by adding a single-step perturbation in the direction of the loss gradient sign, while PGD iteratively applies smaller perturbations and projects the perturbed inputs back to a constrained region around the original inputs after each step. Both attacks were applied to the validation set of Stable Diffusion images using the ResNet50 classifier trained on the same domain.

\begin{table}[h!]
\centering
\caption{Performance Metrics on ResNet50 Model Trained with Stable Diffusion under FGSM and PGD Attacks}
\label{tab:fgsm_pgd_performance}
\begin{tabular}{|l|c|c|c|c|}
\hline
\textbf{Dataset / Attack} & \textbf{Accuracy} & \textbf{Precision} & \textbf{Recall} & \textbf{F1 Score} \\ \hline
Stable Diffusion (No attack)  & 0.9979 & 0.9976 & 0.9982 & 0.9979 \\ \hline
FGSM ($\epsilon$ = 0.01)  & 0.4471 & 0.4699 & 0.8277 & 0.5995 \\ \hline
FGSM ($\epsilon$ = 0.03)  & 0.4820 & 0.4902 & 0.8986 & 0.6343 \\ \hline
FGSM ($\epsilon$ = 0.05)  & 0.5088 & 0.5046 & 0.9525 & 0.6597 \\ \hline
PGD ($\epsilon$ = 0.01)   & 0.3545 & 0.4141 & 0.7011 & 0.5207 \\ \hline
PGD ($\epsilon$ = 0.03)   & 0.4028 & 0.4461 & 0.8047 & 0.5740 \\ \hline
PGD ($\epsilon$ = 0.05)   & 0.4057 & 0.4480 & 0.8115 & 0.5773 \\ \hline
\end{tabular}
\end{table}

\begin{table}[h!]
\centering
\caption{Confusion Matrices for each Attack when tested on ResNet50 model trained with Stable Diffusion (AI=0, Nature=1). TN = AI $\to$ AI, FP = AI $\to$ Nature, FN = Nature $\to$ AI, TP = Nature $\to$ Nature}
\label{tab:conf-matrix-fgsm-pgd}
\begin{tabular}{|l|c|c|c|c|}
\hline
\textbf{Dataset / Attack} & \textbf{TN (0$\to$0)} & \textbf{FP (0$\to$1)} & \textbf{FN (1$\to$0)} & \textbf{TP (1$\to$1)} \\ \hline
Stable Diffusion (No attack)  & 7981 & 19   & 14   & 7986 \\ \hline
FGSM ($\epsilon$ = 0.01)  & 531  & 7469 & 1378 & 6622 \\ \hline
FGSM ($\epsilon$ = 0.03)  & 523  & 7477 & 811  & 7189 \\ \hline
FGSM ($\epsilon$ = 0.05)  & 520  & 7480 & 380  & 7620 \\ \hline
PGD ($\epsilon$ = 0.01)   & 63   & 7937 & 2391 & 5609 \\ \hline
PGD ($\epsilon$ = 0.03)   & 7    & 7993 & 1562 & 6438 \\ \hline
PGD ($\epsilon$ = 0.05)   & 0    & 8000 & 1508 & 6492 \\ \hline
\end{tabular}
\end{table}

\begin{table}[ht!]
\centering
\begin{threeparttable}
\caption{Correctly Predicted‐Accepted (CPA) and Incorrectly Predicted‐Rejected (IPR) under FGSM and PGD adversarial attacks across uncertainty measures.}
\label{tab:adv_attack_cpa_ipr}
\begin{tabular}{llrr}
\toprule
\multirow{2}{*}{\textbf{Dataset}} & \multirow{2}{*}{\textbf{Uncertainty Measure}} 
  & \textbf{CPA} & \textbf{IPR} \\
\cmidrule(lr){3-4}

\midrule
\multirow{8}{*}{\textbf{FGSM $\epsilon=0.01$}} 
 & Total Fisher Information & 97.30 & \textbf{2.63} \\
 & Fisher Frobenius Norm    & 97.32 & 2.77 \\
 & Fisher Entropy           & 97.62 & 9.59 \\
 & Gaussian Process         & 94.80 & \textbf{50.84} \\
 & Entropy of Expected      & \textbf{100.00} & 46.86 \\
 & Knowledge Uncertainty    & \textbf{100.00} & 41.98 \\
 & Combined Uncertainty     & 91.70 & 42.94 \\
\midrule

\multirow{8}{*}{\textbf{FGSM $\epsilon=0.03$}} 
 & Total Fisher Information & 95.02 & \textbf{3.73} \\
 & Fisher Frobenius Norm    & 95.11 & 3.74 \\
 & Fisher Entropy           & 95.54 & 9.00 \\
 & Gaussian Process         & 92.92 & \textbf{67.07} \\
 & Entropy of Expected      & \textbf{100.00} & 46.15 \\
 & Knowledge Uncertainty    & \textbf{100.00} & 46.01 \\
 & Combined Uncertainty     & 91.84 & 52.97 \\
\midrule

\multirow{8}{*}{\textbf{FGSM $\epsilon=0.05$}} 
 & Total Fisher Information & 93.81 & 3.33 \\
 & Fisher Frobenius Norm    & 93.96 & 3.31 \\
 & Fisher Entropy           & 95.48 & 4.95 \\
 & Gaussian Process         & 91.34 & \textbf{65.08} \\
 & Entropy of Expected      & \textbf{100.00} & 48.99 \\
 & Knowledge Uncertainty    & \textbf{100.00} & 51.29 \\
 & Combined Uncertainty     & 91.78 & 50.68 \\
\midrule

\multirow{8}{*}{\textbf{PGD $\epsilon=0.01$}} 
 & Total Fisher Information & 0.00 & 100.00 \\
 & Fisher Frobenius Norm    & 0.00 & 100.00 \\
 & Fisher Entropy           & 100.00 & 0.02 \\
 & Gaussian Process         & 69.96 & \textbf{99.73} \\
 & Entropy of Expected      & \textbf{100}.00 & 27.79 \\
 & Knowledge Uncertainty    & \textbf{100.00} & 24.50 \\
 & Combined Uncertainty     & 93.74 & 74.72 \\
\midrule

\multirow{8}{*}{\textbf{PGD $\epsilon=0.03$}} 
 & Total Fisher Information & 0.00 & \textbf{100.00} \\
 & Fisher Frobenius Norm    & 0.00 & \textbf{100.00} \\
 & Fisher Entropy           & 0.00 & \textbf{100.00} \\
 & Gaussian Process         & 78.21 & \textbf{100.00} \\
 & Entropy of Expected      & \textbf{100.00} & 28.31 \\
 & Knowledge Uncertainty    & \textbf{100.00} & 24.75 \\
 & Combined Uncertainty     & 92.74 & 73.10 \\
\midrule

\multirow{8}{*}{\textbf{PGD $\epsilon=0.05$}} 
 & Total Fisher Information & 0.00 & \textbf{100.00} \\
 & Fisher Frobenius Norm    & 0.00 & \textbf{100.00} \\
 & Fisher Entropy           & 99.98 & 0.00 \\
 & Gaussian Process         & 77.97 & \textbf{100.00} \\
 & Entropy of Expected      & \textbf{100.00} & 28.46 \\
 & Knowledge Uncertainty    & \textbf{100.00} & 25.05 \\
 & Combined Uncertainty     & 91.23 & 74.03 \\
\bottomrule
\end{tabular}
\end{threeparttable}
\end{table}

All inputs were normalised using standard ImageNet statistics before attack generation. Each RGB channel was standardized by subtracting the mean $\mu = [0.485,\ 0.456,\ 0.406]$ and dividing by the standard deviation $\sigma = [0.229,\ 0.224,\ 0.225]$, resulting in normalized inputs of the form:

\begin{equation}
    \text{Input}_{\text{norm}} = \frac{\text{Input} - \mu}{\sigma}
\end{equation}

To ensure perturbed inputs remained valid within the normalised pixel range, we applied clamping after each perturbation using:

\begin{equation}
    \text{Input}_{\text{clamped}} = \min\left(\max\left(\text{Input}_{\text{adv}}, \text{CLAMP}_{\text{MIN}}\right), \text{CLAMP}_{\text{MAX}}\right)
\end{equation}

where $\text{CLAMP}_{\text{MIN}} = \frac{0.0 - \mu}{\sigma}$, $\text{CLAMP}_{\text{MAX}} = \frac{1.0 - \mu}{\sigma}$


The attacks were applied under varying perturbation budgets of $\varepsilon \in {0.01, 0.03, 0.05}$. For PGD, we adopted a step size of $\alpha = \varepsilon / 3$ and performed ten iterations per sample. The impact of these perturbations on the model's classification performance is summarised in Table~\ref{tab:fgsm_pgd_performance}, with corresponding confusion matrices presented in Table~\ref{tab:conf-matrix-fgsm-pgd}.
Under clean, unperturbed conditions, the model achieved near-perfect results, with an accuracy and F1 score of 0.9979. However, even minimal perturbations introduced by adversarial attacks resulted in a substantial performance drop. FGSM with $\varepsilon = 0.01$ reduced accuracy to 44.71\%, with an F1 score of 0.5995. As the perturbation increased to 0.05, the F1 score improved modestly to 0.6597, driven primarily by an increase in recall rather than balanced precision. This trend suggests that under attack, the model increasingly predicts the "Nature" class, correctly identifying most Nature images but at the cost of misclassifying a large number of AI-generated samples. Specifically, precision remains low across all FGSM perturbation levels (approximately 47–50\%), indicating a high false positive rate where AI images are incorrectly classified as Nature.
PGD, being a stronger and iterative attack method, caused even more pronounced degradation. At $\varepsilon = 0.01$, the model's accuracy dropped to 35.45\% with an F1 score of 0.5207. At higher perturbations, performance plateaued at low levels, with accuracy around 40.57\% and an F1 score of 0.5773 at $\varepsilon = 0.05$. These results shows PGD's superior ability to generate effective adversarial examples that exploit the model's vulnerabilities more thoroughly than single-step attacks like FGSM.
The confusion matrices provide further insight into the classification behaviour under attack. Without perturbation, the model correctly classified nearly all examples, with only 19 false positives and 14 false negatives. However, under FGSM, the number of true negatives (correctly identified AI samples) dropped dramatically—from 531 at $\varepsilon = 0.01$ to just 520 at $\varepsilon = 0.05$ while false positives surged to over 7400. This indicates that the model's decision boundary becomes heavily skewed, favouring Nature predictions across the board. Similarly, PGD attacks essentially eliminated the model's ability to detect AI-generated content: at $\varepsilon = 0.05$, the true negative count fell to zero, meaning every AI image was misclassified as Nature. While the model continued to correctly identify a large number of Nature images (e.g., 6492 true positives), this came at the cost of complete failure to reject adversarial AI samples.
These findings highlight a significant asymmetry in the model's robustness. Under adversarial conditions, particularly with PGD, the classifier becomes overly confident in assigning the Nature label, likely because the perturbations push the input representations into regions of the feature space associated with natural images. This behaviour is especially concerning for a detector designed to flag AI-generated content, as high false positive rates mean that synthetic images go undetected, potentially undermining trust and safety applications.

\begin{figure}
    \centering
    \includegraphics[width=0.6\linewidth]{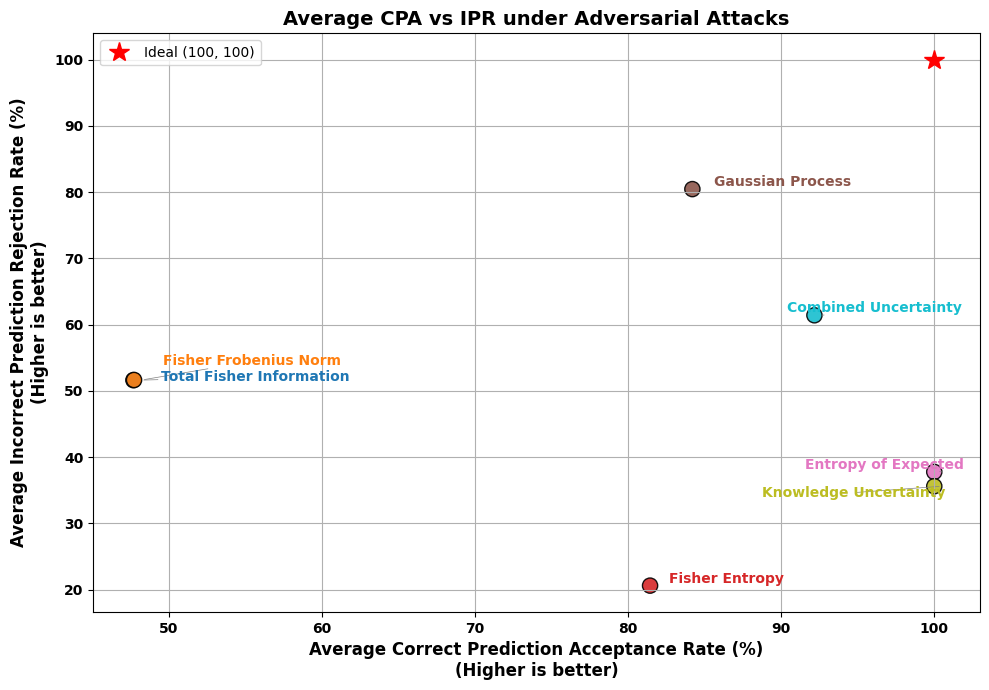}
    \caption{Trade-off between average Correct Prediction Acceptance (CPA) and Incorrect Prediction Rejection (IPR), evaluated by averaging their values over all adversarial attack types across the range of perturbation magnitudes ($\epsilon$) employed in the experiment.}
    \label{fig:adverserial}
\end{figure}

To enhance robustness against adversarial attacks, we evaluated an uncertainty-based rejection mechanism. We tested the implemented uncertainty measures, namely Fisher Information-based metrics, Gaussian Process (GP) variance, Monte Carlo Dropout–based metrics, and a Combined Uncertainty approach, under both FGSM and PGD attacks at $\varepsilon \in {0.01, 0.03, 0.05}$. The detailed CPA (Correctly Predicted–Accepted) and IPR (Incorrectly Predicted–Rejected) results are reported in Table \ref{tab:adv_attack_cpa_ipr} for each perturbation level.
For a clearer assessment, we computed the average CPA and IPR across all perturbation levels and both attack types, capturing each method’s overall reliability, as illustrated in Figure~\ref{fig:adverserial}.
Among all methods, Gaussian Process uncertainty achieved the highest IPR (80.45\%), meaning it was the most effective at rejecting incorrect predictions, a key goal in adversarial settings. It also maintained a solid CPA of 84.20\%, showing that it retained a good portion of correct predictions despite its aggressive filtering.
The Combined Uncertainty method also performed well, averaging 61.41\% IPR and a higher CPA of 92.17\%. While its IPR was slightly lower than GP’s, it struck a more balanced trade-off, making it a strong general-purpose option, especially when both reliability and usability are priorities.
Fisher Entropy offered the best balance among the Fisher-based metrics, with 81.44\% CPA and 20.59\% IPR, providing selective but more conservative rejection. In contrast, Total Fisher Information and Frobenius Norm were overly cautious under stronger attacks, averaging only around 47.7\% CPA and 51.6\% IPR, making them less suitable for robust rejection.
Monte Carlo Dropout–based metrics like Entropy of Expected had perfect CPA (100.00\%) but only 37.76\% IPR, meaning they preserved all correct predictions but missed many incorrect ones, limiting their effectiveness in adversarial defence.

In summary, when prioritising high rejection of incorrect predictions with reasonable retention of correct ones, Gaussian Process uncertainty proves most effective. Combined Uncertainty offers a practical alternative, delivering moderate performance in both acceptance of correct prediction and rejection of incorrect prediction.

\begin{figure}
    \centering
    \includegraphics[width=0.6\linewidth]{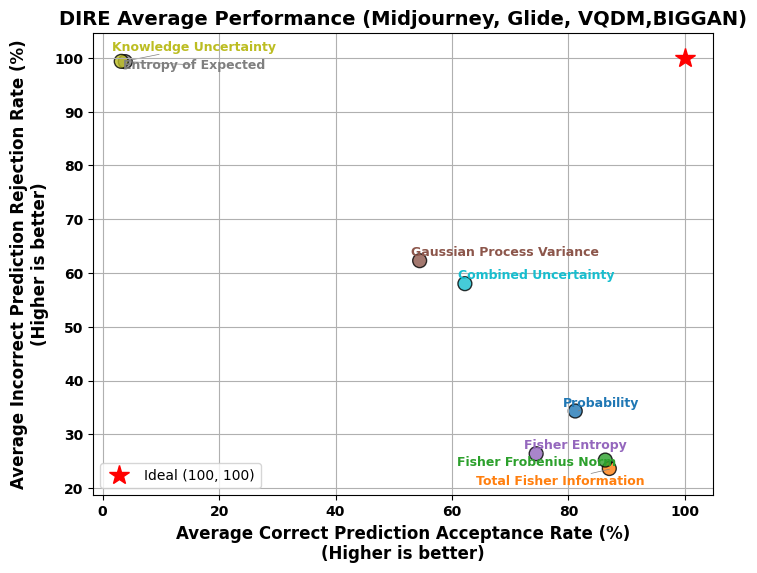}
    \caption{Trade-off between average Correct Prediction Acceptance (CPA) and Incorrect Prediction Rejection (IPR), computed as the mean across the Midjourney, GLIDE, VQDM, and BigGAN datasets using the DIRE model.}
    \label{fig:DIRE}
\end{figure}

\subsubsection*{Comparison with Other Related Work}
\begin{table}[h!]
\centering
\caption{Comparison of Performance Metrics across Datasets for FreqNet, DIRE, GLCM+CNN, LBP+CNN, and Our classifier (AI = 0, Nature = 1), trained on Stable Diffusion}
\label{tab:stacked_performance_metrics}
\begin{tabular}{|l|l|c|c|c|c|}
\hline
\textbf{Dataset} & \textbf{Model} & \textbf{Accuracy} & \textbf{Precision} & \textbf{Recall} & \textbf{F1 Score} \\ \hline

\multirow{5}{*}{Stable Diffusion} 
    & FreqNet & 0.9889 & 0.9861 & 0.9918 & 0.9889 \\
    & DIRE & 0.9521 & 0.9151 & 0.9968 & 0.9542 \\
    & GLCM+CNN & 0.9863 & 0.9815 & 0.9902 & 0.9858 \\
    & LBP+CNN & 0.9788 & 0.9724 & 0.9845 & 0.9784 \\
    & Our classifier & \textbf{0.9979} & \textbf{0.9976} & \textbf{0.9982} & \textbf{0.9979} \\ \hline

\multirow{5}{*}{Midjourney} 
    & FreqNet & 0.5381 & 0.5200 & 0.9915 & 0.6822 \\
    & DIRE & 0.5490 & 0.5261 & 0.9860 & 0.6862 \\
    & GLCM+CNN & 0.6525 & 0.6397 & 0.6651 & 0.6522 \\
    & LBP+CNN & 0.5985 & 0.5905 & 0.6065 & 0.5981 \\
    & Our classifier & \textbf{0.6602} & \textbf{0.5957} & \textbf{0.9972} & \textbf{0.7458} \\ \hline

\multirow{5}{*}{GLIDE} 
    & FreqNet & 0.5158 & 0.5080 & 0.9947 & 0.6726 \\
    & DIRE & 0.5385 & 0.5203 & 0.9875 & 0.6815 \\
    & GLCM+CNN & 0.6649 & 0.6521 & 0.6741 & 0.6629 \\
    & LBP+CNN & 0.6341 & 0.6521 & 0.7018 & 0.6795 \\
    & Our classifier & \textbf{0.7361} & \textbf{0.6550} & \textbf{0.9977} & \textbf{0.7908} \\ \hline

\multirow{5}{*}{VQDM} 
    & FreqNet & 0.6487 & 0.5880 & 0.9933 & 0.7387 \\
    & DIRE & 0.5788 & 0.5434 & 0.9873 & 0.7010 \\
    & GLCM+CNN & 0.6795 & 0.6668 & 0.6889 & 0.6777 \\
    & LBP+CNN & 0.6239 & 0.6102 & 0.6324 & 0.6218 \\
    & Our classifier & \textbf{0.5717} & \textbf{0.5387} & \textbf{0.9965} & \textbf{0.6994} \\ \hline

\multirow{5}{*}{BigGAN} 
    & FreqNet & 0.5980 & 0.5548 & 0.9922 & 0.7117 \\
    & DIRE & 0.5262 & 0.5137 & 0.9875 & 0.6758 \\
    & GLCM+CNN & 0.6690 & 0.6613 & 0.6809 & 0.6709 \\
    & LBP+CNN & 0.6149 & 0.6052 & 0.6264 & 0.6155 \\
    & Our classifier & \textbf{0.5597} & \textbf{0.5319} & \textbf{0.9975} & \textbf{0.6938} \\ \hline

\end{tabular}
\end{table}

The core innovation of our approach lies in its ability to reject incorrect predictions under out-of-distribution (OOD) conditions rather than relying solely on raw classification accuracy. Unlike conventional classifiers that provide deterministic outputs for every input, our method introduces a rejection mechanism based on uncertainty, allowing the model to defer decisions when confidence is low, a capability that becomes crucial under dataset shift.
Prior methods, such as FreqNet \cite{tan2024frequency} and, DIRE offer interpretable, feature-based predictions but lack this flexibility. FreqNet detects spectral anomalies believed to be induced by architectural artefacts (e.g., upsampling layers) in GAN-generated images. DIRE\cite{wang2023dire}, in contrast, leverages reconstruction residuals from a pre-trained diffusion model, operating under the assumption that synthetic images are better reconstructed than natural ones due to their proximity to the learned generation manifold. While both models offer intuitive insights, they produce fixed outputs for all samples, regardless of confidence, making them vulnerable when exposed to unseen data from generators like Midjourney, GLIDE, VQDM, or BigGAN.

To explore alternative representations, we also implemented two texture-based hybrid approaches, GLCM+CNN and LBP+CNN, that combine handcrafted feature extraction with deep learning. The GLCM+CNN pipeline computes Gray-Level Co-occurrence Matrices to capture spatial texture patterns and passes them to a CNN for classification. This method has shown promise in tasks like presentation attack detection~\cite{glcmcnn1} and histopathology classification~\cite{glcmcnn2}. Similarly, LBP+CNN encodes local texture variations using Local Binary Patterns, which are then fed into a CNN. LBP-based features have proven effective in liveness detection and synthetic face recognition~\cite{lbpcnn} due to their robustness to subtle visual artefacts.
When evaluated on our cross-generator detection task, both hybrid models performed well on in-distribution data (Stable Diffusion), achieving F1-scores of 0.9863 (GLCM+CNN) and 0.9788 (LBP+CNN). However, their generalisation of OOD data declined significantly. For example, GLCM+CNN reached only 0.6522 on Midjourney and 0.6629 on GLIDE, while LBP+CNN dropped to 0.5981 and 0.6795 on the same datasets. These results suggest that while handcrafted texture features capture certain class-specific artefacts, they are less adaptive to shifts across generative models.
In comparison, our ResNet50 model trained on Stable Diffusion achieves better generalisation across all OOD datasets. As shown in Tables~\ref{tab:stacked_performance_metrics} and~\ref{tab:cross_resnet_performance}, ResNet50 outperforms FreqNet, DIRE, and the hybrid models on Midjourney (F1 = 0.7458) and GLIDE (F1 = 0.7908). While FreqNet performs slightly better on VQDM (0.7387 vs 0.6994) and GLCM+CNN approaches similar performance (0.6777), our model demonstrates more consistent cross-generator robustness.
More importantly, our approach does not stop at classification. The critical distinction is in our confidence-aware rejection mechanism, which enables the model to handle OOD inputs more safely. By combining uncertainty estimates from Fisher Information, Gaussian Process variance, and Monte Carlo Dropout, the system can flag and reject low-confidence predictions. This not only improves reliability but also prevents overconfident misclassifications without requiring retraining for each new data distribution. In high-risk applications where misclassified synthetic content must be blocked, this rejection capability is far more impactful than accuracy alone.

Importantly, we demonstrate that this rejection mechanism is model-agnostic. To validate this, we trained a separate ResNet on DIRE’s residual features and confirmed that all uncertainty measures from our main pipeline remain applicable. As shown in Figure~\ref{fig:DIRE}, the trade-off between average Correct Prediction Acceptance (CPA) and Incorrect Prediction Rejection (IPR) remains consistent, underscoring the robustness of our approach. This illustrates that our rejection strategy is not limited to deep pixel-space classifiers; it can also be effectively integrated into interpretable, feature-driven models such as DIRE, FreqNet, or hybrid architectures built from handcrafted features like GLCM+CNN and LBP+CNN.

\section*{Conclusion}
The empirical results across both in-distribution (Stable Diffusion) and shifted datasets (Midjourney, Glide, VQDM, BigGAN and StyleGAN3) highlight the critical role of uncertainty measures in controlling the acceptance or rejection of samples under varying data distributions. When the test data closely resembles the training distribution, standard metrics such as probability, total Fisher information, and Fisher Frobenius norm excel; they accept nearly all correct predictions while successfully rejecting most incorrect ones. However, under dataset shift, particularly involving AI-generated samples, their effectiveness at rejecting incorrect predictions diminishes significantly, indicating a vulnerability to unseen or evolving generative methods.

Conversely, highly conservative measures such as Entropy of Expected and Knowledge Uncertainty tend to reject nearly all newly encountered AI data, resulting in near-perfect incorrect rejection rates. However, this aggressive behaviour comes at a substantial cost: a large portion of correct predictions across both the AI and Nature classes are also discarded. In contrast, the Combined Uncertainty measure demonstrates a more balanced and robust performance under dataset shift. While it does reject more correct out-of-distribution AI predictions than standard measures, it consistently achieves a high incorrect rejection rate of approximately 70\% on previously unseen generative models. Importantly, it still retains a reasonable proportion of correct predictions, particularly for the Nature class and in-distribution AI images, where rejection is minimal.
In high-risk scenarios, this conservative approach is often desirable to adopt a stricter stance on AI predictions, as it helps quarantine potentially misclassified synthetic samples. These rejected, but high-quality examples can then be leveraged during retraining, enabling the model to better adapt to evolving generative distributions.
This trade-off between aggressiveness and selectivity is also observed under adversarial attacks. When subjected to FGSM and PGD perturbations, the Gaussian Process-based uncertainty achieves approximately 80\% rejection of successful adversarial examples, outperforming other individual measures in terms of error filtering. However, it also exhibits a lower acceptance rate for correct predictions, retaining only about 85\% of them. In contrast, the Combined Uncertainty measure rejects around 60\% of adversarially perturbed errors while preserving over 90\% of correct predictions. It is a more reliable and conservative defence mechanism that balances caution with accuracy.

In summary, this analysis underscores that as dataset shift intensifies, whether through novel generative models or more pronounced domain shifts, no single measure perfectly preserves the balance between correct acceptance and incorrect rejection. Retraining the model or adopting advanced domain adaptation strategies may be necessary to maintain stringent performance standards. Nevertheless, the Combined Uncertainty strategy presented here offers a versatile framework that can be tuned according to specific risk tolerances. By leveraging multiple uncertainty indicators together, it mitigates blind spots inherent in individual measures, supporting more informed decision-making when confronted with rapidly evolving synthetic image generation methods.

\section*{Data availability}
The datasets used in this study are publicly available, ensuring transparency and reproducibility of our results. Specifically, we utilized the GenImage dataset, which can be accessed from the GenImage repository at \url{https://github.com/GenImage-Dataset/GenImage}, and the related paper link is: \url{https://arxiv.org/abs/2306.08571}. In addition, we incorporated the StyleGAN3-generated face dataset, available on Kaggle at \url{https://www.kaggle.com/datasets/mayankjha146025/fake-face-images-generated-from-different-gans}
, which we used as a supplementary out-of-distribution benchmark. To complement these synthetic faces, we employed real face images from the Kaggle “Deepfake Face Images” dataset \url{https://www.kaggle.com/datasets/kshitizbhargava/deepfake-face-images}
, ensuring a balanced evaluation between genuine and StyleGAN3-generated samples.

\section*{Acknowledgment}
The authors from Northumbria University are thankful for the funding for this project from the Alan Turing Institute, UK, with grant number DSGCAIS-100009.




 





\bibliography{sample}

@article{zhu2024genimage,
  author = {Zhu, M. and Chen, H. and Yan, Q. and Huang, X. and Lin, G. and Li, W. and Tu, Z. and Hu, H. and Hu, J. and Wang, Y.},
  title = {GenImage: A million-scale benchmark for detecting AI-generated images},
  journal = {Advances in Neural Information Processing Systems},
  volume = {36},
  year = {2024}
}

@misc{durall,
  author = {Durall, R. and Keuper, M. and Pfreundt, F.-J. and Keuper, J.},
  title = {Unmasking DeepFakes with simple Features},
  note = {arXiv:1911.00686 [cs, stat]},
  year = {2020},
  url = {https://arxiv.org/abs/1911.00686}
}

@inproceedings{bradshaw,
  author = {Bradshaw, J. and Matthews, A. and Ghahramani, Z.},
  title = {Adversarial Examples, Uncertainty, and Transfer Testing Robustness in Gaussian Processes and Neural Networks},
  booktitle = {International Conference on Artificial Intelligence and Statistics (AISTATS)},
  pages = {233-244},
  year = {2017}
}

@misc{brock2018large,
  author = {Brock, A.},
  title = {Large Scale GAN Training for High Fidelity Natural Image Synthesis},
  note = {ArXiv Preprint ArXiv:1809.11096},
  year = {2018}
}

@misc{nichol2021glide,
  author = {Nichol, A. and Dhariwal, P. and Ramesh, A. and Shyam, P. and Mishkin, P. and McGrew, B. and Sutskever, I. and Chen, M.},
  title = {GLIDE: Towards photorealistic image generation and editing with text-guided diffusion models},
  note = {ArXiv Preprint ArXiv:2112.10741},
  year = {2021}
}

@inproceedings{gu2022vector,
  author = {Gu, S. and Chen, D. and Bao, J. and Wen, F. and Zhang, B. and Chen, D. and Yuan, L. and Guo, B.},
  title = {Vector quantized diffusion model for text-to-image synthesis},
  booktitle = {Proceedings of the IEEE/CVF Conference on Computer Vision and Pattern Recognition},
  pages = {10696-10706},
  year = {2022}
}

@inproceedings{rombach2022high,
  author = {Rombach, R. and Blattmann, A. and Lorenz, D. and Esser, P. and Ommer, B.},
  title = {High-resolution image synthesis with latent diffusion models},
  booktitle = {Proceedings of the IEEE/CVF Conference on Computer Vision and Pattern Recognition},
  pages = {10684-10695},
  year = {2022}
}

@article{dhariwal2021diffusion,
  author = {Dhariwal, P. and Nichol, A.},
  title = {Diffusion models beat GANs on image synthesis},
  journal = {Advances in Neural Information Processing Systems},
  volume = {34},
  pages = {8780-8794},
  year = {2021}
}

@misc{midjourney2022,
  author = {Midjourney},
  title = {Midjourney},
  year = {2022},
  note = {Online},
  url = {https://www.midjourney.com/home/}
}

@inproceedings{wang2020cnn,
  title={CNN-generated images are surprisingly easy to spot... for now},
  author={Wang, Sheng-Yu and Wang, Oliver and Zhang, Richard and Owens, Andrew and Efros, Alexei A},
  booktitle={Proceedings of the IEEE/CVF conference on computer vision and pattern recognition},
  pages={8695--8704},
  year={2020}
}

@inproceedings{gragnaniello2021gan,
  title={Are GAN generated images easy to detect? A critical analysis of the state-of-the-art},
  author={Gragnaniello, Diego and Cozzolino, Davide and Marra, Francesco and Poggi, Giovanni and Verdoliva, Luisa},
  booktitle={2021 IEEE international conference on multimedia and expo (ICME)},
  pages={1--6},
  year={2021},
  organization={IEEE}
}

@inproceedings{ju2022fusing,
  title={Fusing global and local features for generalized ai-synthesized image detection},
  author={Ju, Yan and Jia, Shan and Ke, Lipeng and Xue, Hongfei and Nagano, Koki and Lyu, Siwei},
  booktitle={2022 IEEE International Conference on Image Processing (ICIP)},
  pages={3465--3469},
  year={2022},
  organization={IEEE}
}

@inproceedings{sinitsa2024deep,
  title={Deep image fingerprint: Towards low budget synthetic image detection and model lineage analysis},
  author={Sinitsa, Sergey and Fried, Ohad},
  booktitle={Proceedings of the IEEE/CVF Winter Conference on Applications of Computer Vision},
  pages={4067--4076},
  year={2024}
}

@inproceedings{dogoulis2023improving,
  title={Improving synthetically generated image detection in cross-concept settings},
  author={Dogoulis, Pantelis and Kordopatis-Zilos, Giorgos and Kompatsiaris, Ioannis and Papadopoulos, Symeon},
  booktitle={Proceedings of the 2nd ACM International Workshop on Multimedia AI against Disinformation},
  pages={28--35},
  year={2023}
}

@article{cozzolino2018forensictransfer,
  title={Forensictransfer: Weakly-supervised domain adaptation for forgery detection},
  author={Cozzolino, Davide and Thies, Justus and R{\"o}ssler, Andreas and Riess, Christian and Nie{\ss}ner, Matthias and Verdoliva, Luisa},
  journal={arXiv preprint arXiv:1812.02510},
  year={2018}
}

@inproceedings{zhang2019detecting,
  title={Detecting and simulating artifacts in gan fake images},
  author={Zhang, Xu and Karaman, Svebor and Chang, Shih-Fu},
  booktitle={2019 IEEE international workshop on information forensics and security (WIFS)},
  pages={1--6},
  year={2019},
  organization={IEEE}
}

@article{park2024performance,
  title={Performance Comparison and Visualization of AI-Generated-Image Detection Methods},
  author={Park, Daeeol and Na, Hyunsik and Choi, Daeseon},
  journal={IEEE Access},
  year={2024},
  publisher={IEEE}
}

@inproceedings{epstein2023online,
  title={Online detection of ai-generated images},
  author={Epstein, David C and Jain, Ishan and Wang, Oliver and Zhang, Richard},
  booktitle={Proceedings of the IEEE/CVF International Conference on Computer Vision},
  pages={382--392},
  year={2023}
}

@inproceedings{gal2016dropout,
  title={Dropout as a bayesian approximation: Representing model uncertainty in deep learning},
  author={Gal, Yarin and Ghahramani, Zoubin},
  booktitle={international conference on machine learning},
  pages={1050--1059},
  year={2016},
  organization={PMLR}
}

@article{wang2019aleatoric,
  title={Aleatoric uncertainty estimation with test-time augmentation for medical image segmentation with convolutional neural networks},
  author={Wang, Guotai and Li, Wenqi and Aertsen, Michael and Deprest, Jan and Ourselin, S{\'e}bastien and Vercauteren, Tom},
  journal={Neurocomputing},
  volume={338},
  pages={34--45},
  year={2019},
  publisher={Elsevier}
}

@article{filos2019systematic,
  title={A systematic comparison of bayesian deep learning robustness in diabetic retinopathy tasks},
  author={Filos, Angelos and Farquhar, Sebastian and Gomez, Aidan N and Rudner, Tim GJ and Kenton, Zachary and Smith, Lewis and Alizadeh, Milad and De Kroon, Arnoud and Gal, Yarin},
  journal={arXiv preprint arXiv:1912.10481},
  year={2019}
}

@article{gour2022uncertainty,
  title={Uncertainty-aware convolutional neural network for COVID-19 X-ray images classification},
  author={Gour, Mahesh and Jain, Sweta},
  journal={Computers in biology and medicine},
  volume={140},
  pages={105047},
  year={2022},
  publisher={Elsevier}
}

@inproceedings{deng2023uncertainty,
  title={Uncertainty estimation by fisher information-based evidential deep learning},
  author={Deng, Danruo and Chen, Guangyong and Yu, Yang and Liu, Furui and Heng, Pheng-Ann},
  booktitle={International Conference on Machine Learning},
  pages={7596--7616},
  year={2023},
  organization={PMLR}
}

@article{lee2024radiation,
  title={Radiation image reconstruction and uncertainty quantification using a Gaussian process prior},
  author={Lee, Jaewon and Joshi, Tenzing H and Bandstra, Mark S and Gunter, Donald L and Quiter, Brian J and Cooper, Reynold J and Vetter, Kai},
  journal={Scientific Reports},
  volume={14},
  number={1},
  pages={22958},
  year={2024},
  publisher={Nature Publishing Group UK London}
}

@inproceedings{dutordoir2020bayesian,
  title={Bayesian image classification with deep convolutional Gaussian processes},
  author={Dutordoir, Vincent and Wilk, Mark and Artemev, Artem and Hensman, James},
  booktitle={International Conference on Artificial Intelligence and Statistics},
  pages={1529--1539},
  year={2020},
  organization={PMLR}
}

@article{goodfellow2020generative,
  title={Generative adversarial networks},
  author={Goodfellow, Ian and Pouget-Abadie, Jean and Mirza, Mehdi and Xu, Bing and Warde-Farley, David and Ozair, Sherjil and Courville, Aaron and Bengio, Yoshua},
  journal={Communications of the ACM},
  volume={63},
  number={11},
  pages={139--144},
  year={2020},
  publisher={ACM New York, NY, USA}
}

@inproceedings{lu2018image,
  title={Image generation from sketch constraint using contextual gan},
  author={Lu, Yongyi and Wu, Shangzhe and Tai, Yu-Wing and Tang, Chi-Keung},
  booktitle={Proceedings of the European conference on computer vision (ECCV)},
  pages={205--220},
  year={2018}
}

@article{ho2020denoising,
  title={Denoising diffusion probabilistic models},
  author={Ho, Jonathan and Jain, Ajay and Abbeel, Pieter},
  journal={Advances in neural information processing systems},
  volume={33},
  pages={6840--6851},
  year={2020}
}

@article{hendrycks2016baseline,
  title={A baseline for detecting misclassified and out-of-distribution examples in neural networks},
  author={Hendrycks, Dan and Gimpel, Kevin},
  journal={arXiv preprint arXiv:1610.02136},
  year={2016}
}

@article{liang2017enhancing,
  title={Enhancing the reliability of out-of-distribution image detection in neural networks},
  author={Liang, Shiyu and Li, Yixuan and Srikant, Rayadurgam},
  journal={arXiv preprint arXiv:1706.02690},
  year={2017}
}

@inproceedings{he2016deep,
  title={Deep residual learning for image recognition},
  author={He, Kaiming and Zhang, Xiangyu and Ren, Shaoqing and Sun, Jian},
  booktitle={Proceedings of the IEEE conference on computer vision and pattern recognition},
  pages={770--778},
  year={2016}
}

@inproceedings{glcmcnn1,
  author={M. Akçay, S. Akçay and T. M. Breckon},
  title={Using Deep Convolutional Neural Network Architectures for Face Presentation Attack Detection},
  booktitle={IEEE Transactions on Biometrics},
  year={2019}
}

@article{glcmcnn2,
  author={A. Uçar, A. G. Yalçın and H. Demirel},
  title={Deep Feature Extraction using GLCM and CNN in Histopathology Image Classification},
  journal={IEEE Access},
  year={2024}
}

@inproceedings{lbpcnn,
  author={K. C. Mohan and C. R. Hegde},
  title={Facial Liveness Detection using Local Binary Patterns and Convolutional Neural Networks},
  booktitle={Intelligent Systems and Applications, Springer},
  year={2024}
}

@inproceedings{tan2024frequency,
  title={Frequency-aware deepfake detection: Improving generalizability through frequency space domain learning},
  author={Tan, Chuangchuang and Zhao, Yao and Wei, Shikui and Gu, Guanghua and Liu, Ping and Wei, Yunchao},
  booktitle={Proceedings of the AAAI Conference on Artificial Intelligence},
  volume={38},
  number={5},
  pages={5052--5060},
  year={2024}
}

@inproceedings{wang2023dire,
  title={Dire for diffusion-generated image detection},
  author={Wang, Zhendong and Bao, Jianmin and Zhou, Wengang and Wang, Weilun and Hu, Hezhen and Chen, Hong and Li, Houqiang},
  booktitle={Proceedings of the IEEE/CVF International Conference on Computer Vision},
  pages={22445--22455},
  year={2023}
}

@article{nie2024detecting,
  title={Detecting Discrepancies Between AI-Generated and Natural Images Using Uncertainty},
  author={Nie, Jun and Zhang, Yonggang and Liu, Tongliang and Cheung, Yiu-ming and Han, Bo and Tian, Xinmei},
  journal={arXiv preprint arXiv:2412.05897},
  year={2024}
}

@misc{Jha2022FakeFaceGANs,
  author       = {Mayank Jha},
  title        = {Fake Face Images Generated from Different GANs},
  howpublished = {Kaggle dataset},
  year         = {2022},
  note         = {StyleGAN3 subset used; 10,392 training images, 2,598 test images},
  url          = {https://www.kaggle.com/datasets/mayankjha146025/fake-face-images-generated-from-different-gans}
}

@misc{Bhargava_DeepfakeFaceImages_Real_2022,
  author       = {Kshitiz Bhargava},
  title        = {StyleGan‐StyleGan2 Deepfake Face Images (Real Faces) [Kaggle Dataset]},
  howpublished = {Kaggle},
  year         = {2022},
  note         = {Real face images component; 70,000 real \& 70,000 fake images total},
  url          = {https://www.kaggle.com/datasets/kshitizbhargava/deepfake-face-images}
}





\section*{Author contributions statement}

R.Y., B.I. and N.A. designed and conceived the experiment(s) by looking at the latest research work; R.Y. and E.K.B conducted the experiment(s); R.Y. and B.I. wrote the paper; J.C. and F.K. gave feedback and suggestions to improve the experiments and presentation of results. All authors reviewed the results and analysed the manuscript.

\section*{Additional information}
\textbf{Competing interests}: The author(s) declare no competing interests.





\end{document}